\documentclass{article} 

\usepackage{colm2024_conference}
\UseRawInputEncoding

\usepackage{amsmath,amsfonts,bm}

\def\eqref#1{equation~\ref{#1}}

\def\1{\bm{1}}

\DeclareMathAlphabet{\mathsfit}{\encodingdefault}{\sfdefault}{m}{sl}
\SetMathAlphabet{\mathsfit}{bold}{\encodingdefault}{\sfdefault}{bx}{n}

\usepackage{microtype}
\usepackage{hyperref}
\usepackage{url}
\usepackage{listings}
\usepackage{graphicx}
\usepackage{amsmath}
\usepackage{subcaption}

\usepackage[ruled,noend]{algorithm2e}

\usepackage{lipsum}
\usepackage{epigraph}
\usepackage{booktabs}
\definecolor{darkblue}{rgb}{0, 0, 0.5}

\definecolor{citecolor}{HTML}{2779af}
\definecolor{linkcolor}{HTML}{c0392b}
\hypersetup{colorlinks=true, breaklinks=true, citecolor=citecolor, linkcolor=linkcolor, urlcolor=linkcolor}

\definecolor{darkgreen}{rgb}{0.0, 0.5, 0.0}

\title{Quiet-STaR: Language Models Can Teach Themselves to Think Before Speaking}

\author{Eric Zelikman \\
Stanford University
\And
Georges Harik\\
Notbad AI Inc
\And
Yijia Shao\\
Stanford University
\And
Varuna Jayasiri\\
Notbad AI Inc
\And
Nick Haber\\
Stanford University
\And
Noah D.~Goodman\\
Stanford University
}

\colmfinalcopy 

\begin{document}

\maketitle

\begin{abstract}
When writing and talking, people sometimes pause to think. Although reasoning-focused works have often framed reasoning as a method of answering questions or completing agentic tasks, reasoning is implicit in almost all written text. For example, this applies to the steps not stated between the lines of a proof or to the theory of mind underlying a conversation. In the Self-Taught Reasoner (STaR, Zelikman et al. 2022), useful thinking is learned by inferring rationales from few-shot examples in question-answering and learning from those that lead to a correct answer. This is a highly constrained setting -- ideally, a language model could instead learn to infer unstated rationales in arbitrary text. We present \textbf{Quiet-STaR}, a generalization of STaR in which LMs learn to generate rationales at each token to explain future text, improving their predictions. We address key challenges, including 1) the computational cost of generating continuations, 2) the fact that the LM does not initially know how to generate or use internal thoughts, and 3) the need to predict beyond individual next tokens. To resolve these, we propose a tokenwise parallel sampling algorithm, using learnable tokens indicating a thought's start and end, and an extended teacher-forcing technique. Encouragingly, generated rationales disproportionately help model difficult-to-predict tokens and improve the LM's ability to directly answer difficult questions. In particular, after continued pretraining of an LM on a corpus of internet text with Quiet-STaR, we find zero-shot improvements on GSM8K (5.9\%$\rightarrow$10.9\%) and CommonsenseQA (36.3\%$\rightarrow$47.2\%) and observe a perplexity improvement of difficult tokens in natural text. Crucially, these improvements require no fine-tuning on these tasks. Quiet-STaR marks a step towards LMs that can learn to reason in a more general and scalable way. 
\looseness=-1
\end{abstract}

\epigraph{``Life can only be understood backwards; but it must be lived forwards.''}{--- Søren Kierkegaard}


\section{Introduction}
\label{sec:intro}
Much of the meaning of text is hidden between the lines: without understanding why statements appear in a document, a reader has only a shallow understanding. Moreover, this has been repeatedly shown to be true for LMs as well, in the contexts of tasks ranging from commonsense reasoning to theorem proving to programming \citep{wei_chain_2022,nye2021show,zelikman2022star,zelikman2022parsel,kojima_large_2022}. Reasoning about implications of text to predict later text has consistently been shown to improve LM performance on a variety of tasks, but methods for allowing LMs to learn from their reasoning (e.g., \citealt{zelikman2022star}) have focused on solving individual tasks or predefined sets of tasks (e.g., \citealt{wei_finetuned_2021}). These works rely on carefully curated datasets to provide either specific reasoning tasks or in some cases, the reasoning itself. We instead ask, if reasoning is implicit in all text, why shouldn't we leverage the task of language modeling to teach reasoning?

In particular, the Self-Taught Reasoner (STaR, \citealt{zelikman2022star}) showed that LMs can bootstrap their reasoning ability on question-answering (QA) datasets by sampling rationales to attempt to answer questions, training on rationales if they led to a correct final answer, and then repeating this to iteratively solve more difficult problems. Yet, training from curated QA datasets limits the scale and generalizability of the rationales. QA datasets, especially high-quality ones, require thoughtful curation and will inherently only ever cover a subset of reasoning tasks.
Thus, we extend STaR -- instead of the LM learning to reason on particular tasks like mathematical QA, we train an LM to generate reasoning that helps it infer future text from a large internet text corpus. As a result, we allow the LM to learn from the diverse tasks present in language \citep{weber2021language}. 
This builds on an intuition essential to the current language modeling paradigm, namely, that "language models are unsupervised multitask learners" \citep{radford2019language}.
Thus, as in STaR, we leverage the LM's pre-existing reasoning ability to generate rationales and train the LM on them with a REINFORCE-based reward \citep{williams1992simple}. We refer to this technique as Quiet-STaR, as it can be understood as applying STaR ``quietly'', training the model to think before it speaks.

\begin{figure}
  \centering 
\vspace{-30pt}
\includegraphics[height=5.5cm]{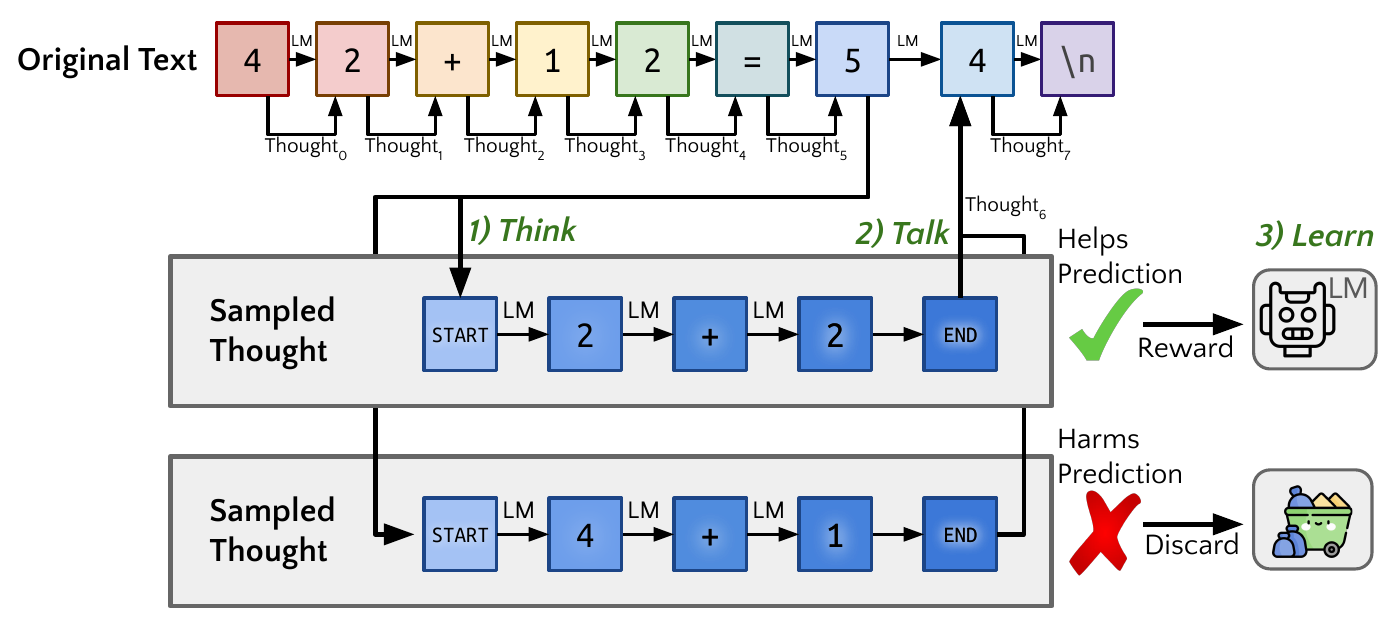} 
\vspace{-2pt}
  \caption{\textbf{Quiet-STaR}. We visualize the algorithm as applied during training to a single thought. We generate thoughts, in parallel, following all tokens in the text \textcolor{darkgreen}{(think)}. The model produces a mixture of its next-token predictions with and without a thought \textcolor{darkgreen}{(talk)}. We apply REINFORCE, as in STaR, to increase the likelihood of thoughts that help the model predict future text while discarding thoughts that make the future text less likely \textcolor{darkgreen}{(learn)}.
  } 
  \vspace{-10px}
  \label{fig:overview}
\end{figure}%

Broadly, Quiet-STaR proceeds by generating rationales after every token to explain future text (\textit{think}), mixing the future-text predictions with and without rationales (\textit{talk}), and then learning to generate better rationales using REINFORCE (\textit{learn}). We apply Quiet-STaR to Mistral 7B \citep{jiang2023mistral} using the web text datasets OpenWebMath \citep{paster2023openwebmath} and Colossal Clean Crawled Corpus (C4, \citealt{raffel2020exploring}).
We find that, even without dataset-specific fine-tuning, Quiet-STaR results in improvements to zero-shot direct-reasoning abilities on CommonsenseQA (36.3\%$\rightarrow$47.2\%) and GSM8K (5.9\%$\rightarrow$10.9\%), and that these improvements consistently increase with the number of tokens used in the LM's internal thoughts. Lastly, we qualitatively investigate patterns in the generated rationales.

In solving this task, we make the following contributions:\vspace{-6px}
\begin{enumerate}\itemsep0em 
    \item We generalize STaR to learn reasoning from diverse unstructured text data. To our knowledge, this is the first work explicitly training LMs to \textbf{reason generally} from text, rather than on curated reasoning tasks or collections of reasoning tasks.
    \item We propose and implement a \textbf{parallel sampling algorithm} that makes our training procedure scalable, generating rationales from all token positions in a given string.
    \item We introduce custom \textbf{meta-tokens} at the start and end of each thought to allow the LM to learn that it should be generating a rationale and when it should make a prediction based on that rationale.
    \item We apply a \textbf{mixing head} to retrospectively determine how much to incorporate the next-token prediction from a given thought into the current next-token prediction.
    \item We show that a \textbf{non-myopic loss}, including multiple tokens ahead for language modeling, improves the effect of thinking.
    \item On multiple tasks, we demonstrate that thinking allows the LM to predict difficult tokens better than one trained on the same web text, improving with longer thoughts.\looseness=-1
\end{enumerate}

\begin{figure}
\vspace{-30px}
\centering
\begin{minipage}{0.49\textwidth}
\centering
\includegraphics[width=1.02\textwidth]{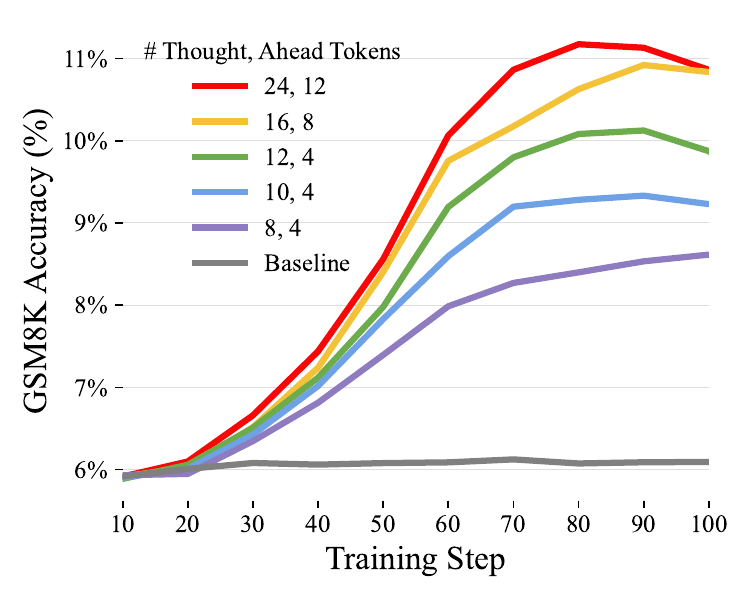}\vspace{-10px}
\caption*{(a) GSM8K}
\end{minipage}\hfill
\begin{minipage}{0.49\textwidth}
\centering
\includegraphics[width=1.02\textwidth]{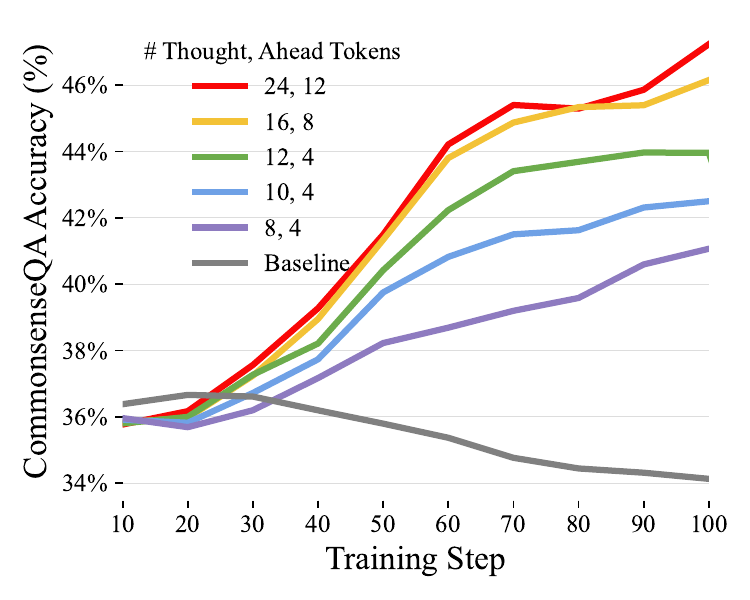}\vspace{-10px}
\caption*{(b) CommonsenseQA}
\end{minipage}
\vspace{-5px}
\caption{\textbf{Generalization Results}. We evaluate the extent to which the model trained with Quiet-STaR generalizes to directly answering problems that require reasoning. The left plot (a) shows the zero-shot accuracy on GSM8K, while the right plot (b) shows the zero-shot accuracy on CommonsenseQA, without any fine-tuning. In both plots, the x-axis represents training steps, and each line corresponds to a different number of thinking tokens used during Quiet-STaR training. The y-axis measures the zero-shot direct accuracy on the respective datasets. We also include an inference normalized version of this plot in Figure~\ref{fig:normalized}.}
\label{fig:generalization}
\vspace{-10px}
\end{figure}

\section{Related Work}
\subsection{Reasoning in Language Models}
There have been many works on training and exploiting language models to solve difficult tasks by first training them to reason through them. For example, \citet{rajani2019explain} demonstrated that a pre-trained language model fine-tuned to output on human reasoning traces before answering multiple-choice commonsense reasoning questions outperformed one trained directly on answers. \citet{shwartz2020unsupervised} demonstrated that language models, when provided with some scaffolding, can generate these helpful chain-of-thought solutions without additional supervision. Later, \citet{nye2021show} demonstrated that ``scratchpads'' required less scaffolding when the language models were more capable, a result later reinforced by \citet{wei_chain_2022}, emphasizing informal tasks, and further strengthened by \citet{kojima_large_2022}, demonstrating this behavior could be accomplished zero-shot. Most recently, \citet{wang2024chain} showed further that for commonsense-question answering, one could force a language model to leverage chain-of-thought reasoning by preventing it from emitting any valid answer tokens unless it was confident. However, once again, these approaches only work for a question-answer dataset, and \citet{wang2024chain} relies on heuristics to identify when the model has output answer tokens. Somewhat like TRICE \citep{phan2023training}, we use the relative improvements in the log-likelihood of the target text across rationales as an estimate of quality, but we simply subtract the mean reward and do not incorporate more complex control variates.

\begin{algorithm}[t]
\caption{Quiet Self-Taught Reasoner (Quiet-STaR)}
\DontPrintSemicolon
\label{alg:qstar}

\KwIn{Language model $\theta_0$, training steps $\mathrm{num\_steps}$, sequence length $l$, thought length $t$, learning rate $\alpha$, batch size $b$, number of thoughts $n_{thoughts}$, number of ground truth tokens used for supervising each thought $n_{true}$}
\KwOut{Language model $\theta$ that generates rationales to predict future text}

\For{$i=0$ \KwTo $\mathrm{num\_steps}$}{
Sample batch of sequences $X$ of length $l$ \\
$h^{init} \gets \mathrm{hidden\_states}_{\theta_{i}}(X)$ \\
\For{$j=1$ \KwTo $l$ \textbf{in parallel using attention mask}}{
{
  $\log p^{\mathrm{init}}_{j:j+n_{true}} \gets \mathrm{lm\_head}_{\theta_{i}}(h_{j:j+n_{true}}^{init})$ \tcp*{Predict next tokens} 
  $T_j \gets \mathrm{generate\_tokens}_{\theta_{i}}([X_{:j}; \texttt{<start\_thought>}], t, n_{thoughts})$  \tcp{Generate thought} 
  $T_j \gets [T_j; \texttt{<end\_thought>}]$ \\
  $h_{j:j+n_{true}}^{\mathrm{thought}} \gets \mathrm{hidden\_states}_{\theta_{i}}([X_{:j}; T_j; X_{j:j + n_{true} - 1}])$ \\
  $\log p_{j:j+n_{true}}^{\mathrm{thought}} \gets \mathrm{lm\_head}_{\theta_{i}}(h_{j:j+n_{true}}^{\mathrm{thought}})$ \tcp*{Predict next tokens w/ thought}
  $w_{j:j+n_{true}} \gets \mathrm{mixing\_head}_{\theta_{i}}(h_{j:j+n_{true}}^{\mathrm{thought}}, h_{j:j+n_{true}}^{init})$ \\
  $\log p_j^{\mathrm{talk}} \gets w_{j:j+n_{true}} \cdot \log p_{j:j+n_{true}}^{\mathrm{init}} + (1 - w_{j:j+n_{true}}) \cdot \log p_{j:j+n_{true}}^{\mathrm{thought}}$ \tcp*{Mix logits}
  $\mathcal{L}_j^{\mathrm{NLL}} \gets -\log p_{j:j+n_{true}}^{\mathrm{talk}}(X_{j+1:j+n_{true} + 1})$ \\
  $r_j = \log p_{j:j+n_{true}}^{\mathrm{talk}}(X_{j+1:j+n_{true} + 1}) - \log \overline{p}_{j:j+n_{true}}^{\mathrm{talk}}(X_{j+1:j+n_{true} + 1})$\\
  $\nabla_\theta \mathcal{L}_j^{\mathrm{REINFORCE}} \gets -r_j\mathbb{1}[r_j > 0] \cdot \nabla_\theta \log p_{\theta_{i}}(T_j | [X_{:j}; \texttt{<start\_thought>}])$ \\
  $\nabla_\theta\mathcal{L}_j \gets \nabla_\theta \mathcal{L}_j^{\mathrm{NLL}} + \nabla_\theta\mathcal{L}_j^{\mathrm{REINFORCE}}$
}
}
$\theta_{i+1} \gets \theta_{i} - \alpha \sum_{j=1}^l \nabla_\theta \mathcal{L}_j$ \tcp*{Update model parameters}
}
\Return{$\theta_{\mathrm{num\_steps}}$}
\end{algorithm}

\subsection{Training Language Models to Reason}
One direction that researchers have used to train language models to reason or improve their reasoning is training the language model on mined reasoning traces or reasoning-like data \citep{rajani2019explain,wei2021finetuned,lewkowycz2022solving,chung2022scaling,gunasekar2023textbooks}. Although this approach has been demonstrated to be effective, it comes with drawbacks. It requires either manual annotation, which is sensitive to the capability of the annotators and is off-policy for the language model (i.e., the distribution of reasoning is not text that the language model would otherwise likely have generated). This approach is also expensive, difficult to scale, and provides no clear path to solving problems harder than those that the annotators are capable of solving.

Another direction for teaching reasoning relies on a language model's own generated reasoning, which can be seen as building on a large body of literature on self-play \citep{silver2017mastering,anthony2017thinking,polu_generative_2020}. These include methods such as the Self-Taught Reasoner \citep{zelikman2022star}, which demonstrated that a language model iteratively trained on its reasoning that led to correct answers could solve increasingly difficult problems. Later work aimed to leverage additional information or assumptions such as \citet{huang2022large} which demonstrated that the algorithm proposed in STaR could still work if one assumed that the majority-vote answer was correct (although this has a lower ultimate performance). Further work has generalized the results of \citet{zelikman2022star}, such as \citet{uesato2022solving} which demonstrated additional usefulness to ``process-based'' supervision where incorrect reasoning traces were filtered, recently V-STaR \citep{hosseini2024v} that demonstrates that training a verifier to guide generation also improves performance, as well as TRICE \citep{hoffman2024training} which maximizes the marginal likelihood of the correct answer given several reasoning traces per problem. Finally, related work has also explored learning intermediate reasoning in the constrained setting of making mathematical statements, where statements in the model's intermediate reasoning could be constrained to only be valid mathematical statements \citep{poesia2023certified}. We include further discussion of related reasoning works in Appendix~\ref{app:otherreasoning}.

\subsection{Meta-tokens}
Recently, a growing body of work has demonstrated the usefulness of custom tokens optimized to perform specific functions in the context of a neural network -- for this reason, they have also been referred to as ``function vectors.'' \citep{todd2023function}. One of the original instantiations of this was prompt-tuning \citep{lester2021power} (and relatedly prefix-tuning \citep{li2021prefix}), where the embeddings corresponding to the tokens of a prompt could be optimized to better accomplish a task. Others have applied meta-tokens to compress long prompts \citep{li2023compressing,jung2023discrete} for efficiency. Most relevant to this work, \citet{mu2024learning} optimized a token such that, when the tokens after it could not attend to the tokens before it (i.e., a context compression token), it would provide sufficient information to future tokens. Although we do not focus on compression, we share the problem of learning a token that affects attention and controls complex downstream behavior. In one related work, \citet{goyal2023think} show that learning a single "pause" token (essentially representing each token as two tokens) improves LM performance. However, unlike the thought tokens in our work, this pause token does not initialize a thought -- instead, it can be seen as acting as the entirety of the thought. We find that reasoning in language is significantly more helpful.

\vspace{-3px}
\section{Problem Statement}\vspace{-3px}
In this work, we introduce an auxiliary `rationale' variable between each pair of  observed tokens of the sequence.
We then aim to optimize a language model with parameters $\theta$ with the capacity to generate intermediate thoughts (or rationales) such that
$$\theta^* = {\arg\max}_\theta E_x \left[ logp_\theta\left(x_{i:n}|x_{0:i},\mathrm{rationale}_\theta\left(x_{0:i}\right)\right) \right]$$
Note that, in principle, this provides no advantage over an optimal language model that already correctly models the language's distribution over strings. Yet, in practice, extensive prior work has shown that language models benefit from intermediate rationales on reasoning tasks \citep{nye2021show,zelikman2022star, wei_chain_2022}. Some work has aimed to explain the effects of chain-of-thought reasoning, namely attributing it to ``locality of experience'' \citep{prystawski2024think}. More broadly, reasoning allows a model to decompose a challenging computation into smaller steps. In effect, we train the model to learn which decomposition and planning steps are effective in predicting future text.
Also note that we formulate the objective as accurately predicting the remaining sequence, rather than only the next token. Once again, for an optimal LM these would be equivalent. However we find that the non-myopic formulation leads to a more effective loss for learning rationales.

\vspace{-3px}
\section{Quiet-STaR}\vspace{-3px}
\subsection{Overview}

Quiet-STaR operates with three main steps (Figure \ref{fig:overview}):\vspace{-5px}
\begin{enumerate}
    \itemsep0em
    \item \textbf{Parallel rationale generation (think, Subsection~\ref{sec:parallel})}: In parallel across $n$ tokens $x_i$ in an input sequence $x_{0:n}$, we generate $r$ rationales of length $t$: $c_i = (c_{i1}, \dots, c_{it})$, resulting in $n \times r$ rationale candidates. We insert learned \lstinline{<|startofthought|>} and \lstinline{<|endofthought|>} tokens to mark each rationale's start and end.
    \item \textbf{Mixing post-rationale and base predictions (talk, Subsection~\ref{sec:mixing})}: From the hidden state output after each rationale, we train a "mixing head" -- a shallow MLP producing a weight determining how much the post-rationale next-token predicted logits should be incorporated compared to the base language model predicted logits. This approach eases distribution shift early in finetuning, due to introducing rationales.\looseness=-1
    \item \textbf{Optimizing rationale generation (learn, Subsection~\ref{sec:optimizingrationales})}: We optimize the rationale generation parameters (start/end tokens and LM weights) to increase the likelihood of rationales that make future text more probable. We use REINFORCE to provide a learning signal to rationales based on their impact on future-token prediction. To reduce variance, we apply a teacher-forcing trick to include in the loss the likelihood of predicting not only the token after the thought but also later tokens.
\end{enumerate}

\begin{figure}
  \centering 
\vspace{-35pt}
\includegraphics[height=6cm]{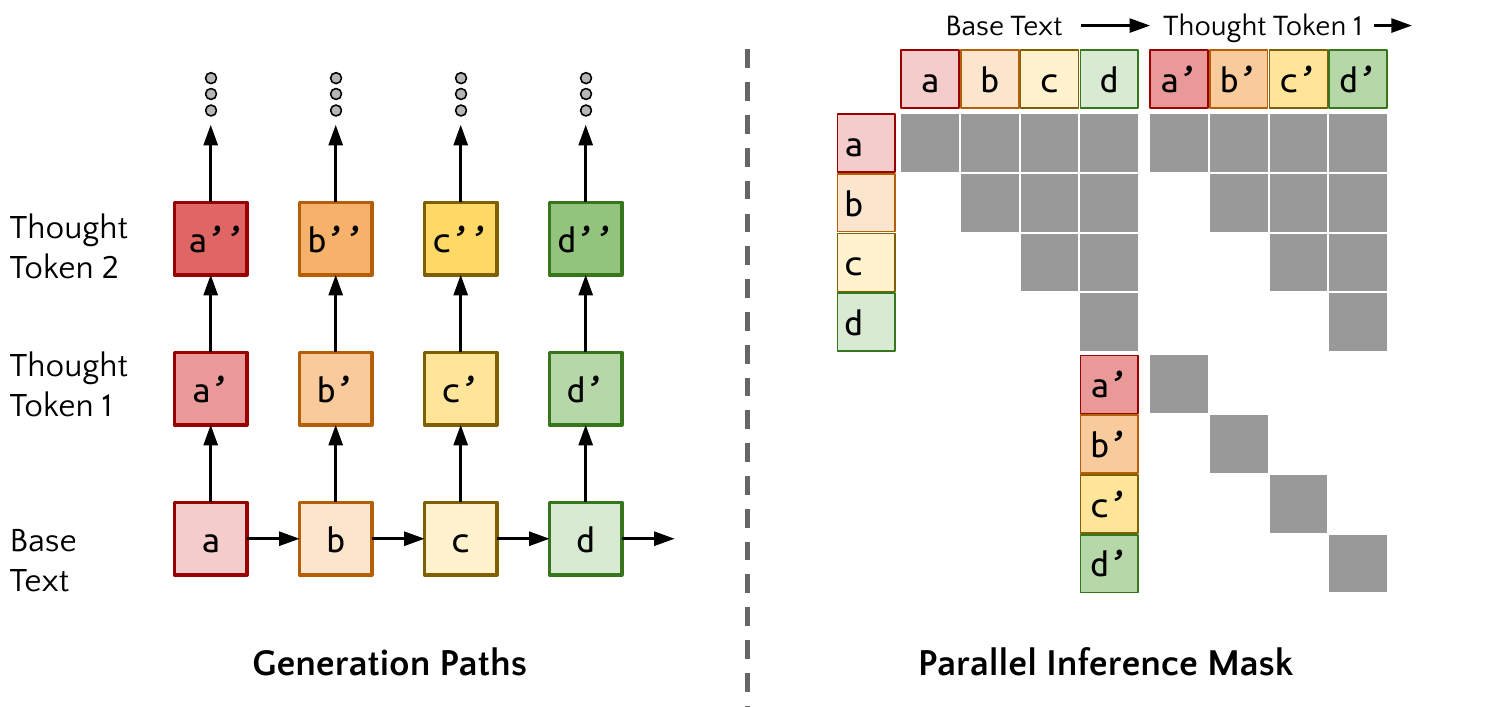}
\vspace{-10pt}
  \caption{\textbf{Parallel Generation}. By constructing an attention mask that allows all thought tokens to pay attention to themselves, all preceding thought tokens within the same thought, and the preceding text, we can generate continuations of all of the thoughts in parallel. Each inference call is used to generate one additional thought token for all text tokens. } 
  \label{fig:parallelgen}
  \vspace{-8px}
\end{figure}%

\subsection{Parallel Generation}
\label{sec:parallel}
A key challenge in Quiet-STaR is efficiently generating rationales at each token position in the input sequence. Naively, this would require a separate forward pass for each token, which becomes computationally intractable for long sequences.

We allow for highly parallel generation by first observing that an inference pass of a language model produces a probability distribution over the next tokens for all input tokens. Naturally, this allows us to sample one next token from each token in the input. If one has generated a successor from each token, it is not possible to simply continue with the original sequence.
For example, imagine predicting the next token after each token of ``$<bos>$ the cat sat'' one might generate ``yes orange saw down'' -- each successor by itself is a reasonable next token to a prefix of the sequence, but the list of tokens is a set of ``counterfactual'' continuations of these prefixes. 
We can, however, leverage these continuations to generate hidden thoughts for each observed token.

To do this efficiently, we cache each forward pass and concatenate a diagonal attention mask to the previous attention mask: each generated token now attends to all of the tokens that were used to generate it, as well as to itself (but not to token on other ``counterfactual'' paths). Moreover, this parallelized next-sampling token procedure can be repeated arbitrarily many times (or at least, until one runs out of memory). We visualize this procedure in Figure~\ref{fig:parallelgen} and highlight additional ways to make this algorithm faster in Appendix~\ref{app:parallelfast}.

\subsection{``Mixing'' (Residual) Heads}\vspace{-3px}
\label{sec:mixing}
When starting with a pre-trained model, thoughts will initially be out of distribution, and hence harm language modeling performance.
To smooth the transition to thinking, 
we introduce a learned interpolation between the LM predictions with and without thoughts. Given the end-of-thought token's hidden state and the hidden state of the original text token, the mixing head outputs a weight that determines the extent to which the post-thought prediction logits will be used. We use a shallow multi-layer perceptron for this head, outputting a scalar for each token. We include implementation details in Appendix~\ref{app:hyperparams}.

\subsection{Optimizing Rationale Generation}
\label{sec:optimizingrationales}

\subsubsection{Optimizing Start-of-Thought and End-of-Thought Tokens}
The \lstinline{<|startofthought|>} and \lstinline{<|endofthought|>} tokens serve as learned meta-tokens that control the model's rationale generation. Optimizing the representation of these tokens, especially the \lstinline{<|startofthought|>} token, is crucial but challenging due to the discrete nature of the rationale tokens.
We initialize the start and end token embeddings to the embedding corresponding to the em dash, "\lstinline{---}", which often appears in text data to denote a pause or thought. This leverages the language model's preexisting knowledge. In addition, to allow these embeddings to be optimized more quickly, we apply a (hyperparameter) weight to the gradients of these embeddings during the update step. Intuitively, the start thought tokens can be understood as putting the model into a ``thinking mode'' and the end thought token can be understood as telling the model when it's done thinking.

\begin{figure}
  \centering 
\vspace{-35pt}
\includegraphics[height=5.5cm]{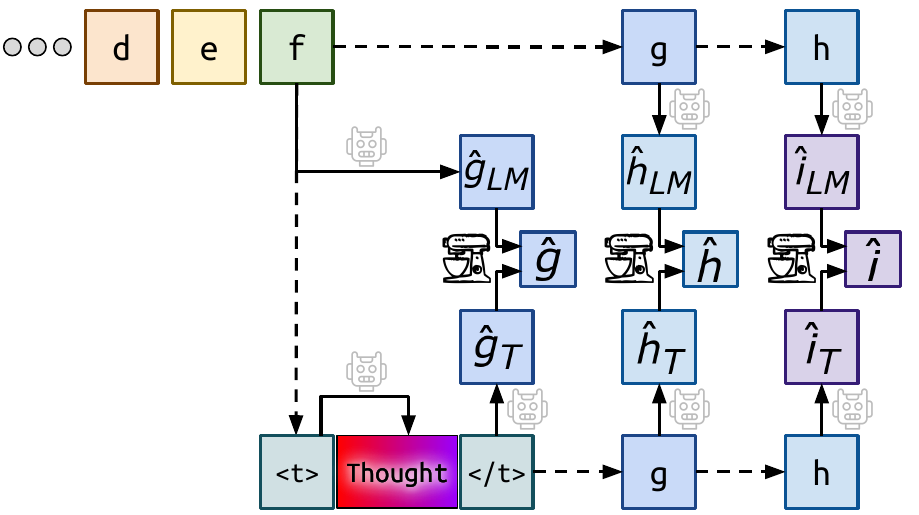}
  \caption{\textbf{Forward Pass and Teacher Forcing}. We visualize a single forward pass of our algorithm. Solid lines denote language model computation, while dashed lines indicate tokens are inserted via teacher forcing, and the mixer represents the mixing head. In particular, we visualize predicting three tokens ahead. Thought generation is shown in more detail in Figure~\ref{fig:overview} and Figure~\ref{fig:parallelgen}.} 
  \vspace{-10pt}
  \label{fig:mechanics}
\end{figure}%

\subsubsection{Non-myopic Scoring and Teacher-forcing}

Because we do not expect thoughts to be useful in predicting every token, 
we would prefer the model's reward to depend less on the exact next word in the text following the thought and more on the following semantic content. There are two primary challenges here. First, unlike in typical language modeling with transformers, only the thoughts corresponding to a given next-token prediction receive a gradient from that prediction---a consequence of our parallel sampling strategy. 
We could address this by adding loss terms for future tokens by sampling the tokens before.
However this would result in much higher entropy for language modeling in general and lower-quality generated text,
because it would train the LM to partially disregard its preceding tokens. Instead, we use the parallel attention mask to compute the log probabilities of the true next tokens, applying teacher forcing by assuming the model selected the correct next ground-truth token (as implicit in normal language modeling with transformers).
Note that the loss for each future token also depends on a mixing weight computed from the end thought token and the previous observed token. The number of future tokens included in the loss is a hyper-parameter.
We apply the same teacher-forcing technique to insert the start and end tokens. We visualize this procedure in Figure~\ref{fig:mechanics}.\looseness=-1

\vspace{-4px}
\subsubsection{Objective}
We use REINFORCE to optimize the likelihoods of the rationales based on their usefullness: the log-likelihood of the $n_{true}$ true next tokens $X_{j+1:j+n_{true} +1}$ under the language model given previous observed tokens and a particular rationale ($p_{j:j+n_{true}}^{\mathrm{talk}}$ as shorthand for the mixed prediction probabilities after thinking, see Algorithm \ref{alg:qstar}).
To reduce variance, we generate multiple rationale continuations for each token in the input sequence (loosely inspired by TRICE, \citet{phan2023training}). We thus define the reward $r_j$ for each rationale $T_j$ as the difference between $p_{j:j+n_{true}}^{\mathrm{talk}}$  and the average across rationales for that token ($\overline{p}_{j:j+n_{true}}^{\mathrm{talk}}$):\vspace{-9px}

$$r_j = \log p_{j:j+n_{true}}^{\mathrm{talk}}(X_{j+1:j+n_{true} +1}) - \log \overline{p}_{j:j+n_{true}}^{\mathrm{talk}}(X_{j + 1:j+n_{true} + 1})$$\vspace{-12px}

We then use this reward in a REINFORCE loss term to update the language model parameters $\theta$ to increase the likelihood of rationales that perform better than the average:\vspace{-10px}

$$\nabla_\theta \mathcal{L}_j^{\mathrm{REINFORCE}} = -r_j \cdot \nabla_\theta \log p_{\theta}(T_j | [X_{:j}; \texttt{<|startofthought|>}])$$\vspace{-15px}

We found it useful to exclude the negative reward from the REINFORCE loss term, as it led to more stable training, though it may introduce some bias.

This loss term encourages the model to generate rationales that improve its predictions of future tokens compared to the average prediction across all generated rationales for that token. The gradients from this loss are used to update both the LM parameters and the start-of-thought and end-of-thought token embeddings, with a (hyperparameter) weight applied to the gradients of the start-of-thought and end-of-thought token embeddings to accelerate their optimization. By iteratively optimizing these parameters, Quiet-STaR trains the model to generate more useful rationales throughout training. Lastly, we also include a log-likelihood loss, $\mathcal{L}_j^{\mathrm{NLL}}$, to ensure that the LM learns to optimize the talking heads and also receives a next-token prediction signal for the base LM head\footnote{Due to our linear mixing, equivalent to shifting the mixing weight toward the base prediction.}.

\vspace{-5px}
\section{Experiments and Results}\vspace{-3px}

 Intuitively, not all tokens require equal amounts of thought. For example, consider the sentence ``the person is run-'': although there is inevitably some probability of the token being something other than ``ing''\footnote{For example, in this very text, the token following ``run'' is ``-''}, as a standalone sentence without context, additional thinking is unlikely to improve a well-trained model's prediction. Indeed, we conjecture that for most chunks of most online text, additional thought has little to no impact. 
 Indeed, early in our exploration we observed that Quiet-STaR does not benefit all tokens equally.
 Thus, we design our experiments to investigate whether our approach is useful in predicting tokens that \textit{do} require thought. We evaluate 1) whether Quiet-STaR improves a language model's ability to directly predict answers in datasets that require reasoning; and, 2) the distribution of impacts resulting from thinking tokens. We conduct all of our experiments starting with the base version of Mistral 7B \citep{jiang2023mistral}. 
 
 We perform most of our experiments by training on OpenWebMath \citep{paster2023openwebmath}, a crawl that emphasizes more technical webpages. We selected OpenWebMath because we anticipated that it would have a higher density of tokens that benefit from reasoning, which our experiments support. We also evaluate Quiet-STaR on C4 \citep{raffel2020exploring}, a widely used LM pretraining corpus with more diverse text, and again show significant albeit smaller benefits.

\subsection{Downstream Performance}
In this subsection, we evaluate the extent to which Quiet-STaR improves the zero-shot reasoning capabilities of the language model on CommonsenseQA \citep{talmor2018commonsenseqa} and GSM8K \citep{cobbe_training_2021}. On CommonsenseQA, we find that Quiet-STaR improves performance by 10.9\% compared to the base language model. As shown in Figure~\ref{fig:generalization}, this improvement consistently increases with the number of tokens used in the model's rationales, indicating that more thorough reasoning through the thought tokens is translating to better direct question-answering performance. Similarly, on GSM8K, Quiet-STaR results in a 5.0\% boost over the base model, and once again, performance scales with the length of the rationales generated during Quiet-STaR training. For reference, in Figure~\ref{fig:generalization}, we include a baseline corresponding to training the same model on the same dataset without thought tokens. 
We observe that in multiple curves, performance appears to eventually deteriorate -- we anticipate that this is because we are not training on these downstream tasks, so the roles of the thought tokens may change over time. We also find a benefit of our non-myopic objective, which we discuss in Appendix~\ref{app:ablations}.

We find that training with Quiet-STaR on C4 \citep{raffel2020exploring} also improves performance on GSM8K ($5.9\% \rightarrow 8.1\%$) and CommonsenseQA ($36.3\% \rightarrow 42.6\%$) but by a smaller margin. Specifically, for our C4 evaluation, we train Mistral 7B with 16 thought tokens and 4 true tokens ahead and otherwise the same setup. 

We can compare these improvements to those offered by pause tokens \citep{goyal2023think}, which can be seen as a constrained version of Quiet-STaR where each token is represented by two tokens and the second "pause" token acts as the entirety of the thought. In particular, our setup is most comparable to their pause token fine-tuning, as we also finetune a pretrained model. Their results indicate that pause token fine-tuning also provides minor gains over the base model on CommonsenseQA, they observed an improvement from 26.9\% to 28.8\%; on GSM8K, \citet{goyal2023think} found that pause token fine-tuning harms performance. Moreover, on both tasks (and the majority of their evaluated tasks), they observed that additional thought tokens harmed performance. Moreover, they discuss the ``lukewarm effect of pause-finetuning a standard-pretrained model'' \citep{goyal2023think}. This suggests that allowing the model to generate multi-token rationales leads to more effective reasoning compared to the single-token "pauses". Note however, that unlike \citet{goyal2023think}, we \textit{do not fine-tune} on the downstream tasks.

Overall, these downstream results validate that training a language model to predict the subtext between the lines of general text data can substantially improve its reasoning capabilities, even on datasets it was not explicitly trained on. The fact that longer rationales consistently lead to better outcomes, and that Quiet-STaR outperforms the constrained pause token approach, supports the notion that Quiet-STaR is successfully teaching the model to leverage its own generated thoughts to reason more thoroughly about the input.

\vspace{-5px}
\subsection{Improvement Distribution}\vspace{-3px}
As visualized in Appendix Figure~\ref{fig:distribution}, we find that on average there is little improvement in the LM's ability to predict arbitrary tokens. But, when we visualize the distribution of relative improvements, there is a disproportionate improvement on more difficult tokens. This reflects the idea that some text tokens are substantially harder and benefit more from careful thought.\looseness=-1

In Appendix Figure~\ref{fig:contribution}, we aim to provide some insight into the kinds of tokens where the improvements occur. Namely, while thinking appears to help for many tokens in the example, inspection suggests it disproportionately help to predict tokens where recalling relevant information is useful, such as the name of an applicable theorem or the start of the next step in a proof. Notably, this would align well with the framing proposed by \citet{prystawski2024think}.

\begin{figure}
  \centering 
\vspace{-35pt}
\includegraphics[height=5.5cm]{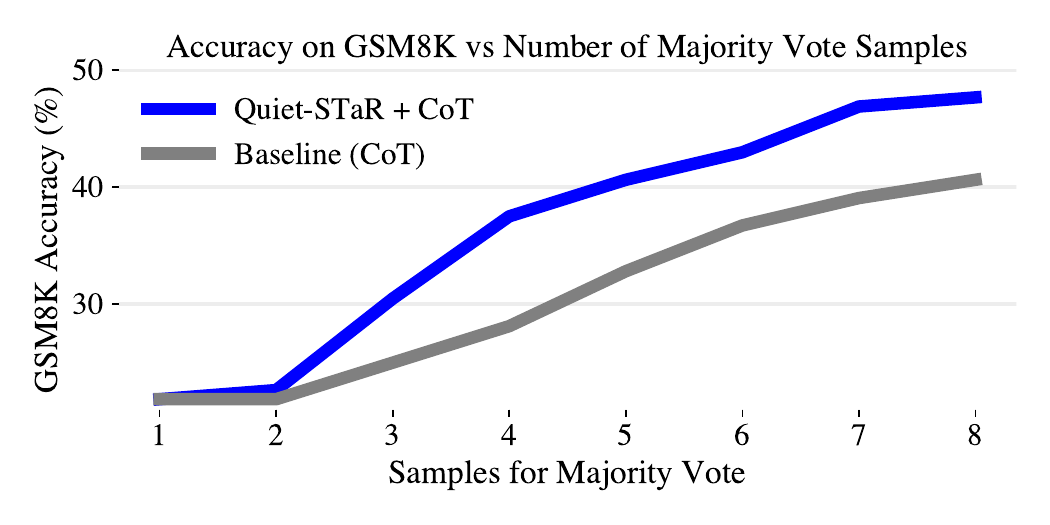}
\vspace{-5pt}
  \caption{\textbf{Zero-shot performance on Quiet-STaR applied to chain-of-thought on GSM8K}. We visualize how using a Quiet-STaR trained Mistral model can improve chain-of-thought performance. We use an 8-thought-token-trained model and use its internal thoughts to improve the tokens in a zero-shot chain-of-thought \citep{kojima_large_2022}} 
  \vspace{-10pt}
  \label{fig:cotplus}
\end{figure}%

\vspace{-3px}
\subsection{Quiet-STaR and Chain-of-Thought}
While there are natural parallels between chain-of-thought prompting and our approach, they are orthogonal and complementary. 
In zero-shot chain-of-thought, a user actively prompts the model to think `out loud', otherwise using its ordinary production distribution \citep{kojima_large_2022}; Quiet-STaR instead allows a model to think quietly at every token, with a distribution trained to be useful. 
We investigate using silent, Quiet-STaR, rationales while generating explicit CoT reasoning. 
Because our goal is generalist reasoning that requires no task-specific input at all, we used a zero-shot prompt (``Let's think step by step.'') without in-context examples.
Our experiments indicate that internal rationales allow the model to generate more structured and coherent chains of thought, shown in Appendix~\ref{app:cotofcot} and visualized in Figure~\ref{fig:cotplus}. 
The majority vote accuracy over 8 samples (cot-maj@8) increases from 40.6\% to 47.7\% with Quiet-STaR, as evaluated on a sample of 128 GSM8K test items. Note that each chain-of-thought solution is sampled with temperature 0.7.\vspace{-3px}


\subsection{Examples}
\lstset{
  basicstyle=\ttfamily,
  columns=fixed,
  fontadjust=true,
  basewidth=0.5em,
  breaklines=True
}

While there is no explicit regularization in Quiet-STaR for thoughts to be human-interpretable, they are generated from the same transformer trained to model language, hence likely to be at least partially understandable. We discuss why this design choice benefits the training stability in Appendix~\ref{instability}.
For reference, we include examples of thoughts generated that were helpful to the model in predicting future tokens in OpenWebMath. First, in one case, recalling that one should start with magnesium to produce magnesium nitride allows it to better predict that the first step of the procedure involves heating magnesium.\looseness=-1

\begin{lstlisting}
'<s> # Magnesium reacts with nitrogen to form magnesium nitride. The chemical formula for this reaction is Mg+N_2-> MgN_2. What is the product, or what are the products, of this reaction?\n\nJan 12, 2016\n\nThe formula for magnesium nitride is $M {g}_{3} {N}_{2}$.\n\n#### Explanation:\n\nAs do many active metals, magnesium nitride can be<|startofthought|> 1 --, so the equation of the reaction that forms magnesium nitride is\n\n$Mg + N_2 \\to<|endofthought|> formed by heating the metal (fier' \end{lstlisting}

In some cases, the most useful thoughts appear to be near-continuations that correspond more closely to the target text, e.g.,

\begin{lstlisting}
An integer $n$ is odd if $n = 2k+1$ for some integer $k$.\n\nTo prove that $A = B$, we must show that $A \\subseteq B$ and $B \\subseteq A$. The first of these tends to<|startthought|> in some sense - to be the more difficult<|endthought|> trickiest for students
\end{lstlisting}

Lastly, we include an example from answering CommonsenseQA. Notably, this thought occurs while reading the question and hence was not used to predict the final answer.
\begin{lstlisting}
'<s> Q: Talking to the same person about the same thing over and over again is<|startofthought|>\n\n(a) a one-to-one correlation\n\n(b) a one-to<|endofthought|> something someone can what?'
\end{lstlisting}

\vspace{-5px}
\section{Limitations}\vspace{-3px}
This work proposes a new framework for learning to reason, and in doing so explores solutions to a variety of meta-learning challenges. However, to solve these challenges, certain simplifications were necessary. For example, it would be valuable to understand whether these techniques work when a model is trained from scratch. 
We have also only applied Quiet-STaR to a 7 billion parameter model, albeit a powerful one. The same techniques applied to a better model would likely yield disproportionately better results, as has often been observed for gains from reasoning \citep{wei_emergent_2022}.

Quiet-STaR results in a substantial overhead, generating many tokens before generating every additional token.
(See Appendix \ref{app:computeadj} for compute adjusted performance results.)
However, this can also be seen as an advantage: typically, a language model can generate the next token based on the current context, and while there are techniques to improve sampling quality, there is no general way to leverage additional compute to enhance next-token prediction. 
In the current implementation we do not support dynamically predicting when to generate, or end, a rationale. However, this would be a natural extension.
For instance, if the mixing head was a prediction from the base language model, before any thought, rather than after the thought, one could apply a threshold to prevent generating thoughts that would not be incorporated. We expect that this is a more difficult task, as predicting the usefulness of a thought is simpler when one has already generated the thought.\looseness=-1

\section{Conclusion}\vspace{-3px}
Quiet-STaR represents a step towards language models that can learn to reason in a general and scalable way. By training on the rich spectrum of reasoning tasks implicit in diverse web text, rather than narrowly specializing for particular datasets, Quiet-STaR points the way to more robust and adaptable language models. Our results demonstrate the promise of this approach, with Quiet-STaR improving downstream reasoning performance while generating qualitatively meaningful rationales. We believe this also opens many potential future directions - for example, one may aim to ensemble thoughts in order to further improve the predictions for future tokens. Moreover, if the language model can predict when thought will be useful, for example by putting the mixing head before the prediction, then the predicted mixing weight could be used to dynamically allocate compute during generation. Future work can build on these insights to further close the gap between language model and human-like reasoning capabilities.

\section*{Ethics Statement}
This work raises some important ethical questions, many of which also apply to STaR. For example, it is impossible to know that the reasoning expressed by the model in language accurately represents the internal processing of the model (i.e., faithfulness). In addition, regardless of faithfulness, there are no safeguards against harmful or biased reasoning patterns if the model finds them useful. Relatedly, we note that CommonsenseQA is known to have many biased questions and low-quality answers \citep{geva2019we}, but we use it in line with prior work \citep{zelikman2022star,goyal2023think}. Thus, aside from improving language modeling, it is unclear in what capacity the rationales themselves should be used.

\section*{Acknowledgements}
We particularly thank Xindi Wu, Michael Li, and Qian Huang for their helpful and detailed comments, as well as Xuechen Li, Jan-Philipp Fr\"anken, Yuhuai Wu, Gabriel Poesia, Winnie Xu, Omar Shaikh, Fan-Yun Sun, Joy He-Yueya, Omar Khattab, and William Yin for useful discussions. In addition, we would like to acknowledge that this work was supported by NSF Grant \#2302701.

\bibliography{colm2024_conference}

\begin{thebibliography}{75}
\providecommand{\natexlab}[1]{#1}
\providecommand{\url}[1]{\texttt{#1}}
\expandafter\ifx\csname urlstyle\endcsname\relax
  \providecommand{\doi}[1]{doi: #1}\else
  \providecommand{\doi}{doi: \begingroup \urlstyle{rm}\Url}\fi

\bibitem[Anthony et~al.(2017)Anthony, Tian, and Barber]{anthony2017thinking}
Thomas Anthony, Zheng Tian, and David Barber.
\newblock Thinking fast and slow with deep learning and tree search.
\newblock \emph{Advances in neural information processing systems}, 30, 2017.

\bibitem[Chen et~al.(2023)Chen, Shu, Shareghi, Collier, Narasimhan, and Yao]{chen2023fireact}
Baian Chen, Chang Shu, Ehsan Shareghi, Nigel Collier, Karthik Narasimhan, and Shunyu Yao.
\newblock Fireact: Toward language agent fine-tuning.
\newblock \emph{arXiv preprint arXiv:2310.05915}, 2023.

\bibitem[Chung et~al.(2022)Chung, Hou, Longpre, Zoph, Tay, Fedus, Li, Wang, Dehghani, Brahma, et~al.]{chung2022scaling}
Hyung~Won Chung, Le~Hou, Shayne Longpre, Barret Zoph, Yi~Tay, William Fedus, Yunxuan Li, Xuezhi Wang, Mostafa Dehghani, Siddhartha Brahma, et~al.
\newblock Scaling instruction-finetuned language models.
\newblock \emph{arXiv preprint arXiv:2210.11416}, 2022.

\bibitem[Cobbe et~al.(2021)Cobbe, Kosaraju, Bavarian, Chen, Jun, Kaiser, Plappert, Tworek, Hilton, Nakano, Hesse, and Schulman]{cobbe_training_2021}
"Karl Cobbe, Vineet Kosaraju, Mohammad Bavarian, Mark Chen, Heewoo Jun, Lukasz Kaiser, Matthias Plappert, Jerry Tworek, Jacob Hilton, Reiichiro Nakano, Christopher Hesse, and John Schulman.
\newblock Training {Verifiers} to {Solve} {Math} {Word} {Problems}.
\newblock \emph{arXiv}, 2021.
\newblock \_eprint: 2110.14168.

\bibitem[Gandhi et~al.(2023)Gandhi, Sadigh, and Goodman]{gandhi2023strategic}
Kanishk Gandhi, Dorsa Sadigh, and Noah~D Goodman.
\newblock Strategic reasoning with language models.
\newblock \emph{arXiv preprint arXiv:2305.19165}, 2023.

\bibitem[Geva et~al.(2019)Geva, Goldberg, and Berant]{geva2019we}
Mor Geva, Yoav Goldberg, and Jonathan Berant.
\newblock Are we modeling the task or the annotator? an investigation of annotator bias in natural language understanding datasets.
\newblock \emph{arXiv preprint arXiv:1908.07898}, 2019.

\bibitem[Goyal et~al.(2023)Goyal, Ji, Rawat, Menon, Kumar, and Nagarajan]{goyal2023think}
Sachin Goyal, Ziwei Ji, Ankit~Singh Rawat, Aditya~Krishna Menon, Sanjiv Kumar, and Vaishnavh Nagarajan.
\newblock Think before you speak: Training language models with pause tokens.
\newblock \emph{arXiv preprint arXiv:2310.02226}, 2023.

\bibitem[Gulcehre et~al.(2023)Gulcehre, Paine, Srinivasan, Konyushkova, Weerts, Sharma, Siddhant, Ahern, Wang, Gu, et~al.]{gulcehre2023reinforced}
Caglar Gulcehre, Tom~Le Paine, Srivatsan Srinivasan, Ksenia Konyushkova, Lotte Weerts, Abhishek Sharma, Aditya Siddhant, Alex Ahern, Miaosen Wang, Chenjie Gu, et~al.
\newblock Reinforced self-training (rest) for language modeling.
\newblock \emph{arXiv preprint arXiv:2308.08998}, 2023.

\bibitem[Gunasekar et~al.(2023)Gunasekar, Zhang, Aneja, Mendes, Del~Giorno, Gopi, Javaheripi, Kauffmann, de~Rosa, Saarikivi, et~al.]{gunasekar2023textbooks}
Suriya Gunasekar, Yi~Zhang, Jyoti Aneja, Caio C{\'e}sar~Teodoro Mendes, Allie Del~Giorno, Sivakanth Gopi, Mojan Javaheripi, Piero Kauffmann, Gustavo de~Rosa, Olli Saarikivi, et~al.
\newblock Textbooks are all you need.
\newblock \emph{arXiv preprint arXiv:2306.11644}, 2023.

\bibitem[Haluptzok et~al.(2023)Haluptzok, Bowers, and Kalai]{haluptzok2023language}
Patrick Haluptzok, Matthew Bowers, and Adam~Tauman Kalai.
\newblock Language models can teach themselves to program better.
\newblock In \emph{The Eleventh International Conference on Learning Representations}, 2023.
\newblock URL \url{https://openreview.net/forum?id=SaRj2ka1XZ3}.

\bibitem[Hewitt et~al.(2023)Hewitt, Thickstun, Manning, and Liang]{hewitt2023backpack}
John Hewitt, John Thickstun, Christopher~D Manning, and Percy Liang.
\newblock Backpack language models.
\newblock \emph{arXiv preprint arXiv:2305.16765}, 2023.

\bibitem[Ho et~al.(2022)Ho, Schmid, and Yun]{ho2022large}
Namgyu Ho, Laura Schmid, and Se-Young Yun.
\newblock Large language models are reasoning teachers.
\newblock \emph{arXiv preprint arXiv:2212.10071}, 2022.

\bibitem[Hoffman et~al.(2024)Hoffman, Phan, Dohan, Douglas, Le, Parisi, Sountsov, Sutton, Vikram, and A~Saurous]{hoffman2024training}
Matthew~Douglas Hoffman, Du~Phan, David Dohan, Sholto Douglas, Tuan~Anh Le, Aaron Parisi, Pavel Sountsov, Charles Sutton, Sharad Vikram, and Rif A~Saurous.
\newblock Training chain-of-thought via latent-variable inference.
\newblock \emph{Advances in Neural Information Processing Systems}, 36, 2024.

\bibitem[Hosseini et~al.(2024)Hosseini, Yuan, Malkin, Courville, Sordoni, and Agarwal]{hosseini2024v}
Arian Hosseini, Xingdi Yuan, Nikolay Malkin, Aaron Courville, Alessandro Sordoni, and Rishabh Agarwal.
\newblock V-star: Training verifiers for self-taught reasoners.
\newblock \emph{arXiv preprint arXiv:2402.06457}, 2024.

\bibitem[Hsieh et~al.(2023)Hsieh, Li, Yeh, Nakhost, Fujii, Ratner, Krishna, Lee, and Pfister]{hsieh2023distilling}
Cheng-Yu Hsieh, Chun-Liang Li, Chih-Kuan Yeh, Hootan Nakhost, Yasuhisa Fujii, Alexander Ratner, Ranjay Krishna, Chen-Yu Lee, and Tomas Pfister.
\newblock Distilling step-by-step! outperforming larger language models with less training data and smaller model sizes.
\newblock \emph{arXiv preprint arXiv:2305.02301}, 2023.

\bibitem[Huang et~al.(2022)Huang, Gu, Hou, Wu, Wang, Yu, and Han]{huang2022large}
Jiaxin Huang, Shixiang~Shane Gu, Le~Hou, Yuexin Wu, Xuezhi Wang, Hongkun Yu, and Jiawei Han.
\newblock Large language models can self-improve.
\newblock \emph{arXiv preprint arXiv:2210.11610}, 2022.

\bibitem[Jang et~al.(2016)Jang, Gu, and Poole]{jang2016categorical}
Eric Jang, Shixiang Gu, and Ben Poole.
\newblock Categorical reparameterization with gumbel-softmax.
\newblock \emph{arXiv preprint arXiv:1611.01144}, 2016.

\bibitem[Jiang et~al.(2023)Jiang, Sablayrolles, Mensch, Bamford, Chaplot, Casas, Bressand, Lengyel, Lample, Saulnier, et~al.]{jiang2023mistral}
Albert~Q Jiang, Alexandre Sablayrolles, Arthur Mensch, Chris Bamford, Devendra~Singh Chaplot, Diego de~las Casas, Florian Bressand, Gianna Lengyel, Guillaume Lample, Lucile Saulnier, et~al.
\newblock Mistral 7b.
\newblock \emph{arXiv preprint arXiv:2310.06825}, 2023.

\bibitem[Jung \& Kim(2023)Jung and Kim]{jung2023discrete}
Hoyoun Jung and Kyung-Joong Kim.
\newblock Discrete prompt compression with reinforcement learning.
\newblock \emph{arXiv preprint arXiv:2308.08758}, 2023.

\bibitem[Khattab et~al.(2022)Khattab, Santhanam, Li, Hall, Liang, Potts, and Zaharia]{khattab2022demonstrate}
Omar Khattab, Keshav Santhanam, Xiang~Lisa Li, David Hall, Percy Liang, Christopher Potts, and Matei Zaharia.
\newblock Demonstrate-search-predict: Composing retrieval and language models for knowledge-intensive nlp.
\newblock \emph{arXiv preprint arXiv:2212.14024}, 2022.

\bibitem[Khattab et~al.(2023)Khattab, Singhvi, Maheshwari, Zhang, Santhanam, Vardhamanan, Haq, Sharma, Joshi, Moazam, et~al.]{khattab2023dspy}
Omar Khattab, Arnav Singhvi, Paridhi Maheshwari, Zhiyuan Zhang, Keshav Santhanam, Sri Vardhamanan, Saiful Haq, Ashutosh Sharma, Thomas~T Joshi, Hanna Moazam, et~al.
\newblock Dspy: Compiling declarative language model calls into self-improving pipelines.
\newblock \emph{arXiv preprint arXiv:2310.03714}, 2023.

\bibitem[Kojima et~al.(2022)Kojima, Gu, Reid, Matsuo, and Iwasawa]{kojima_large_2022}
Takeshi Kojima, Shixiang~Shane Gu, Machel Reid, Yutaka Matsuo, and Yusuke Iwasawa.
\newblock Large {Language} {Models} are {Zero}-{Shot} {Reasoners}, 2022.
\newblock URL \url{https://arxiv.org/abs/2205.11916}.

\bibitem[Lampinen et~al.(2022)Lampinen, Dasgupta, Chan, Matthewson, Tessler, Creswell, McClelland, Wang, and Hill]{lampinen2022can}
Andrew~K Lampinen, Ishita Dasgupta, Stephanie~CY Chan, Kory Matthewson, Michael~Henry Tessler, Antonia Creswell, James~L McClelland, Jane~X Wang, and Felix Hill.
\newblock Can language models learn from explanations in context?
\newblock \emph{arXiv preprint arXiv:2204.02329}, 2022.

\bibitem[Lanchantin et~al.(2024)Lanchantin, Toshniwal, Weston, Sukhbaatar, et~al.]{lanchantin2024learning}
Jack Lanchantin, Shubham Toshniwal, Jason Weston, Sainbayar Sukhbaatar, et~al.
\newblock Learning to reason and memorize with self-notes.
\newblock \emph{Advances in Neural Information Processing Systems}, 36, 2024.

\bibitem[Lester et~al.(2021)Lester, Al-Rfou, and Constant]{lester2021power}
Brian Lester, Rami Al-Rfou, and Noah Constant.
\newblock The power of scale for parameter-efficient prompt tuning.
\newblock \emph{arXiv preprint arXiv:2104.08691}, 2021.

\bibitem[Lewkowycz et~al.(2022)Lewkowycz, Andreassen, Dohan, Dyer, Michalewski, Ramasesh, Slone, Anil, Schlag, Gutman-Solo, et~al.]{lewkowycz2022solving}
Aitor Lewkowycz, Anders Andreassen, David Dohan, Ethan Dyer, Henryk Michalewski, Vinay Ramasesh, Ambrose Slone, Cem Anil, Imanol Schlag, Theo Gutman-Solo, et~al.
\newblock Solving quantitative reasoning problems with language models.
\newblock \emph{Advances in Neural Information Processing Systems}, 35:\penalty0 3843--3857, 2022.

\bibitem[Li et~al.(2024)Li, Fox, and Goodman]{li2024automated}
Michael~Y Li, Emily~B Fox, and Noah~D Goodman.
\newblock Automated statistical model discovery with language models.
\newblock \emph{arXiv preprint arXiv:2402.17879}, 2024.

\bibitem[Li et~al.(2022)Li, Chen, Shen, Chen, Zhang, Li, Wang, Qian, Peng, Mao, et~al.]{li2022explanations}
Shiyang Li, Jianshu Chen, Yelong Shen, Zhiyu Chen, Xinlu Zhang, Zekun Li, Hong Wang, Jing Qian, Baolin Peng, Yi~Mao, et~al.
\newblock Explanations from large language models make small reasoners better.
\newblock \emph{arXiv preprint arXiv:2210.06726}, 2022.

\bibitem[Li \& Liang(2021)Li and Liang]{li2021prefix}
Xiang~Lisa Li and Percy Liang.
\newblock Prefix-tuning: Optimizing continuous prompts for generation.
\newblock \emph{arXiv preprint arXiv:2101.00190}, 2021.

\bibitem[Li et~al.(2023)Li, Dong, Lin, and Guerin]{li2023compressing}
Yucheng Li, Bo~Dong, Chenghua Lin, and Frank Guerin.
\newblock Compressing context to enhance inference efficiency of large language models.
\newblock \emph{arXiv preprint arXiv:2310.06201}, 2023.

\bibitem[Liu et~al.(2023)Liu, Pasunuru, Hajishirzi, Choi, and Celikyilmaz]{liu2023crystal}
Jiacheng Liu, Ramakanth Pasunuru, Hannaneh Hajishirzi, Yejin Choi, and Asli Celikyilmaz.
\newblock Crystal: Introspective reasoners reinforced with self-feedback.
\newblock \emph{arXiv preprint arXiv:2310.04921}, 2023.

\bibitem[Luo et~al.(2023)Luo, Sun, Xu, Zhao, Lou, Tao, Geng, Lin, Chen, and Zhang]{luo2023wizardmath}
Haipeng Luo, Qingfeng Sun, Can Xu, Pu~Zhao, Jianguang Lou, Chongyang Tao, Xiubo Geng, Qingwei Lin, Shifeng Chen, and Dongmei Zhang.
\newblock Wizardmath: Empowering mathematical reasoning for large language models via reinforced evol-instruct.
\newblock \emph{arXiv preprint arXiv:2308.09583}, 2023.

\bibitem[Madaan et~al.(2023)Madaan, Tandon, Gupta, Hallinan, Gao, Wiegreffe, Alon, Dziri, Prabhumoye, Yang, et~al.]{madaan2023self}
Aman Madaan, Niket Tandon, Prakhar Gupta, Skyler Hallinan, Luyu Gao, Sarah Wiegreffe, Uri Alon, Nouha Dziri, Shrimai Prabhumoye, Yiming Yang, et~al.
\newblock Self-refine: Iterative refinement with self.
\newblock \emph{Feedback}, 2023.

\bibitem[Mnih et~al.(2013)Mnih, Kavukcuoglu, Silver, Graves, Antonoglou, Wierstra, and Riedmiller]{mnih2013playing}
Volodymyr Mnih, Koray Kavukcuoglu, David Silver, Alex Graves, Ioannis Antonoglou, Daan Wierstra, and Martin Riedmiller.
\newblock Playing atari with deep reinforcement learning.
\newblock \emph{arXiv preprint arXiv:1312.5602}, 2013.

\bibitem[Mnih et~al.(2016)Mnih, Badia, Mirza, Graves, Lillicrap, Harley, Silver, and Kavukcuoglu]{mnih2016asynchronous}
Volodymyr Mnih, Adria~Puigdomenech Badia, Mehdi Mirza, Alex Graves, Timothy Lillicrap, Tim Harley, David Silver, and Koray Kavukcuoglu.
\newblock Asynchronous methods for deep reinforcement learning.
\newblock In \emph{International conference on machine learning}, pp.\  1928--1937. PMLR, 2016.

\bibitem[Mu et~al.(2024)Mu, Li, and Goodman]{mu2024learning}
Jesse Mu, Xiang Li, and Noah Goodman.
\newblock Learning to compress prompts with gist tokens.
\newblock \emph{Advances in Neural Information Processing Systems}, 36, 2024.

\bibitem[Nye et~al.(2021)Nye, Andreassen, Gur-Ari, Michalewski, Austin, Bieber, Dohan, Lewkowycz, Bosma, Luan, et~al.]{nye2021show}
Maxwell Nye, Anders~Johan Andreassen, Guy Gur-Ari, Henryk Michalewski, Jacob Austin, David Bieber, David Dohan, Aitor Lewkowycz, Maarten Bosma, David Luan, et~al.
\newblock Show your work: Scratchpads for intermediate computation with language models.
\newblock \emph{arXiv preprint arXiv:2112.00114}, 2021.

\bibitem[Pan et~al.(2024)Pan, Jones, Jagadeesan, and Steinhardt]{pan2024feedback}
Alexander Pan, Erik Jones, Meena Jagadeesan, and Jacob Steinhardt.
\newblock Feedback loops with language models drive in-context reward hacking.
\newblock \emph{arXiv preprint arXiv:2402.06627}, 2024.

\bibitem[Paster et~al.(2023)Paster, Santos, Azerbayev, and Ba]{paster2023openwebmath}
Keiran Paster, Marco~Dos Santos, Zhangir Azerbayev, and Jimmy Ba.
\newblock Openwebmath: An open dataset of high-quality mathematical web text.
\newblock \emph{arXiv preprint arXiv:2310.06786}, 2023.

\bibitem[Phan et~al.(2023)Phan, Hoffman, Douglas, Le, Parisi, Sountsov, Sutton, Vikram, Saurous, et~al.]{phan2023training}
Du~Phan, Matthew~Douglas Hoffman, Sholto Douglas, Tuan~Anh Le, Aaron~T Parisi, Pavel Sountsov, Charles Sutton, Sharad Vikram, Rif~A Saurous, et~al.
\newblock Training chain-of-thought via latent-variable inference.
\newblock In \emph{Thirty-seventh Conference on Neural Information Processing Systems}, 2023.

\bibitem[Poesia et~al.(2023)Poesia, Gandhi, Zelikman, and Goodman]{poesia2023certified}
Gabriel Poesia, Kanishk Gandhi, Eric Zelikman, and Noah~D Goodman.
\newblock Certified reasoning with language models.
\newblock \emph{arXiv preprint arXiv:2306.04031}, 2023.

\bibitem[Polu \& Sutskever(2020)Polu and Sutskever]{polu_generative_2020}
Stanislas Polu and Ilya Sutskever.
\newblock Generative {Language} {Modeling} for {Automated} {Theorem} {Proving}.
\newblock \emph{CoRR}, abs/2009.03393, 2020.
\newblock URL \url{https://arxiv.org/abs/2009.03393}.
\newblock \_eprint: 2009.03393.

\bibitem[Prystawski et~al.(2024)Prystawski, Li, and Goodman]{prystawski2024think}
Ben Prystawski, Michael Li, and Noah Goodman.
\newblock Why think step by step? reasoning emerges from the locality of experience.
\newblock \emph{Advances in Neural Information Processing Systems}, 36, 2024.

\bibitem[Qiao et~al.(2024)Qiao, Zhang, Fang, Luo, Zhou, Jiang, Lv, and Chen]{qiao2024autoact}
Shuofei Qiao, Ningyu Zhang, Runnan Fang, Yujie Luo, Wangchunshu Zhou, Yuchen~Eleanor Jiang, Chengfei Lv, and Huajun Chen.
\newblock Autoact: Automatic agent learning from scratch via self-planning.
\newblock \emph{arXiv preprint arXiv:2401.05268}, 2024.

\bibitem[Qiu et~al.(2023)Qiu, Jiang, Lu, Sclar, Pyatkin, Bhagavatula, Wang, Kim, Choi, Dziri, et~al.]{qiu2023phenomenal}
Linlu Qiu, Liwei Jiang, Ximing Lu, Melanie Sclar, Valentina Pyatkin, Chandra Bhagavatula, Bailin Wang, Yoon Kim, Yejin Choi, Nouha Dziri, et~al.
\newblock Phenomenal yet puzzling: Testing inductive reasoning capabilities of language models with hypothesis refinement.
\newblock \emph{arXiv preprint arXiv:2310.08559}, 2023.

\bibitem[Radford et~al.(2019)Radford, Wu, Child, Luan, Amodei, Sutskever, et~al.]{radford2019language}
Alec Radford, Jeffrey Wu, Rewon Child, David Luan, Dario Amodei, Ilya Sutskever, et~al.
\newblock Language models are unsupervised multitask learners.
\newblock \emph{OpenAI blog}, 1\penalty0 (8):\penalty0 9, 2019.

\bibitem[Raffel et~al.(2020)Raffel, Shazeer, Roberts, Lee, Narang, Matena, Zhou, Li, and Liu]{raffel2020exploring}
Colin Raffel, Noam Shazeer, Adam Roberts, Katherine Lee, Sharan Narang, Michael Matena, Yanqi Zhou, Wei Li, and Peter~J Liu.
\newblock Exploring the limits of transfer learning with a unified text-to-text transformer.
\newblock \emph{Journal of machine learning research}, 21\penalty0 (140):\penalty0 1--67, 2020.

\bibitem[Rajani et~al.(2019)Rajani, McCann, Xiong, and Socher]{rajani2019explain}
Nazneen~Fatema Rajani, Bryan McCann, Caiming Xiong, and Richard Socher.
\newblock Explain yourself! leveraging language models for commonsense reasoning.
\newblock In \emph{Proceedings of the 57th Annual Meeting of the Association for Computational Linguistics}, pp.\  4932--4942, 2019.

\bibitem[Schick et~al.(2024)Schick, Dwivedi-Yu, Dess{\`\i}, Raileanu, Lomeli, Hambro, Zettlemoyer, Cancedda, and Scialom]{schick2024toolformer}
Timo Schick, Jane Dwivedi-Yu, Roberto Dess{\`\i}, Roberta Raileanu, Maria Lomeli, Eric Hambro, Luke Zettlemoyer, Nicola Cancedda, and Thomas Scialom.
\newblock Toolformer: Language models can teach themselves to use tools.
\newblock \emph{Advances in Neural Information Processing Systems}, 36, 2024.

\bibitem[Schulman et~al.(2017)Schulman, Wolski, Dhariwal, Radford, and Klimov]{schulman2017proximal}
John Schulman, Filip Wolski, Prafulla Dhariwal, Alec Radford, and Oleg Klimov.
\newblock Proximal policy optimization algorithms.
\newblock \emph{arXiv preprint arXiv:1707.06347}, 2017.

\bibitem[Schuster et~al.(2021)Schuster, Kalyan, Polozov, and Kalai]{schuster_programming_2021}
Tal Schuster, Ashwin Kalyan, Alex Polozov, and Adam~Tauman Kalai.
\newblock Programming {Puzzles}.
\newblock In \emph{Thirty-fifth {Conference} on {Neural} {Information} {Processing} {Systems}}, 2021.
\newblock URL \url{https://openreview.net/forum?id=fe_hCc4RBrg}.

\bibitem[Shinn et~al.(2023)Shinn, Cassano, Labash, Gopinath, Narasimhan, and Yao]{shinn2023reflexion}
Noah Shinn, Federico Cassano, Beck Labash, Ashwin Gopinath, Karthik Narasimhan, and Shunyu Yao.
\newblock Reflexion: Language agents with verbal reinforcement learning.
\newblock \emph{arXiv preprint arXiv:2303.11366}, 2023.

\bibitem[Shwartz et~al.(2020)Shwartz, West, Le~Bras, Bhagavatula, and Choi]{shwartz2020unsupervised}
Vered Shwartz, Peter West, Ronan Le~Bras, Chandra Bhagavatula, and Yejin Choi.
\newblock Unsupervised commonsense question answering with self-talk.
\newblock In \emph{Proceedings of the 2020 Conference on Empirical Methods in Natural Language Processing (EMNLP)}, pp.\  4615--4629, 2020.

\bibitem[Silver et~al.(2017)Silver, Hubert, Schrittwieser, Antonoglou, Lai, Guez, Lanctot, Sifre, Kumaran, Graepel, et~al.]{silver2017mastering}
David Silver, Thomas Hubert, Julian Schrittwieser, Ioannis Antonoglou, Matthew Lai, Arthur Guez, Marc Lanctot, Laurent Sifre, Dharshan Kumaran, Thore Graepel, et~al.
\newblock Mastering chess and shogi by self-play with a general reinforcement learning algorithm.
\newblock \emph{arXiv preprint arXiv:1712.01815}, 2017.

\bibitem[Talmor et~al.(2018)Talmor, Herzig, Lourie, and Berant]{talmor2018commonsenseqa}
Alon Talmor, Jonathan Herzig, Nicholas Lourie, and Jonathan Berant.
\newblock Commonsenseqa: A question answering challenge targeting commonsense knowledge.
\newblock \emph{arXiv preprint arXiv:1811.00937}, 2018.

\bibitem[Todd et~al.(2023)Todd, Li, Sharma, Mueller, Wallace, and Bau]{todd2023function}
Eric Todd, Millicent~L Li, Arnab~Sen Sharma, Aaron Mueller, Byron~C Wallace, and David Bau.
\newblock Function vectors in large language models.
\newblock \emph{arXiv preprint arXiv:2310.15213}, 2023.

\bibitem[Uesato et~al.(2022)Uesato, Kushman, Kumar, Song, Siegel, Wang, Creswell, Irving, and Higgins]{uesato2022solving}
Jonathan Uesato, Nate Kushman, Ramana Kumar, Francis Song, Noah Siegel, Lisa Wang, Antonia Creswell, Geoffrey Irving, and Irina Higgins.
\newblock Solving math word problems with process-and outcome-based feedback.
\newblock \emph{Neural Information Processing Systems (NeurIPS 2022) Workshop on MATH-AI}, 2022.

\bibitem[Wang et~al.(2023)Wang, Zelikman, Poesia, Pu, Haber, and Goodman]{wang2023hypothesis}
Ruocheng Wang, Eric Zelikman, Gabriel Poesia, Yewen Pu, Nick Haber, and Noah~D Goodman.
\newblock Hypothesis search: Inductive reasoning with language models.
\newblock \emph{arXiv preprint arXiv:2309.05660}, 2023.

\bibitem[Wang \& Zhou(2024)Wang and Zhou]{wang2024chain}
Xuezhi Wang and Denny Zhou.
\newblock Chain-of-thought reasoning without prompting.
\newblock \emph{arXiv preprint arXiv:2402.10200}, 2024.

\bibitem[Weber et~al.(2021)Weber, Jumelet, Bruni, and Hupkes]{weber2021language}
Lucas Weber, Jaap Jumelet, Elia Bruni, and Dieuwke Hupkes.
\newblock Language modelling as a multi-task problem.
\newblock \emph{arXiv preprint arXiv:2101.11287}, 2021.

\bibitem[Wei et~al.(2021{\natexlab{a}})Wei, Bosma, Zhao, Guu, Yu, Lester, Du, Dai, and Le]{wei2021finetuned}
Jason Wei, Maarten Bosma, Vincent Zhao, Kelvin Guu, Adams~Wei Yu, Brian Lester, Nan Du, Andrew~M Dai, and Quoc~V Le.
\newblock Finetuned language models are zero-shot learners.
\newblock In \emph{International Conference on Learning Representations}, 2021{\natexlab{a}}.

\bibitem[Wei et~al.(2021{\natexlab{b}})Wei, Bosma, Zhao, Guu, Yu, Lester, Du, Dai, and Le]{wei_finetuned_2021}
Jason Wei, Maarten Bosma, Vincent~Y Zhao, Kelvin Guu, Adams~Wei Yu, Brian Lester, Nan Du, Andrew~M Dai, and Quoc~V Le.
\newblock Finetuned language models are zero-shot learners.
\newblock \emph{arXiv preprint arXiv:2109.01652}, 2021{\natexlab{b}}.

\bibitem[Wei et~al.(2022{\natexlab{a}})Wei, Tay, Bommasani, Raffel, Zoph, Borgeaud, Yogatama, Bosma, Zhou, Metzler, Chi, Hashimoto, Vinyals, Liang, Dean, and Fedus]{wei_emergent_2022}
Jason Wei, Yi~Tay, Rishi Bommasani, Colin Raffel, Barret Zoph, Sebastian Borgeaud, Dani Yogatama, Maarten Bosma, Denny Zhou, Donald Metzler, Ed~H. Chi, Tatsunori Hashimoto, Oriol Vinyals, Percy Liang, Jeff Dean, and William Fedus.
\newblock Emergent {Abilities} of {Large} {Language} {Models}, October 2022{\natexlab{a}}.
\newblock URL \url{http://arxiv.org/abs/2206.07682}.
\newblock arXiv:2206.07682 [cs].

\bibitem[Wei et~al.(2022{\natexlab{b}})Wei, Wang, Schuurmans, Bosma, Ichter, Xia, Chi, Le, and Zhou]{wei_chain_2022}
Jason Wei, Xuezhi Wang, Dale Schuurmans, Maarten Bosma, Brian Ichter, Fei Xia, Ed~Chi, Quoc Le, and Denny Zhou.
\newblock Chain of {Thought} {Prompting} {Elicits} {Reasoning} in {Large} {Language} {Models}, 2022{\natexlab{b}}.
\newblock URL \url{https://arxiv.org/abs/2201.11903}.

\bibitem[Williams(1992)]{williams1992simple}
Ronald~J Williams.
\newblock Simple statistical gradient-following algorithms for connectionist reinforcement learning.
\newblock \emph{Machine learning}, 8:\penalty0 229--256, 1992.

\bibitem[Yao et~al.(2022)Yao, Zhao, Yu, Du, Shafran, Narasimhan, and Cao]{yao2022react}
Shunyu Yao, Jeffrey Zhao, Dian Yu, Nan Du, Izhak Shafran, Karthik Narasimhan, and Yuan Cao.
\newblock React: Synergizing reasoning and acting in language models.
\newblock \emph{International Conference on Learning Representations (ICLR 2023)}, 2022.

\bibitem[Zelikman et~al.(2022)Zelikman, Wu, Mu, and Goodman]{zelikman2022star}
Eric Zelikman, Yuhuai Wu, Jesse Mu, and Noah Goodman.
\newblock Star: Bootstrapping reasoning with reasoning.
\newblock \emph{Advances in Neural Information Processing Systems}, 35:\penalty0 15476--15488, 2022.

\bibitem[Zelikman et~al.(2023{\natexlab{a}})Zelikman, Huang, Poesia, Goodman, and Haber]{zelikman2022parsel}
Eric Zelikman, Qian Huang, Gabriel Poesia, Noah~D. Goodman, and Nick Haber.
\newblock Parsel: Algorithmic reasoning with language models by composing decompositions, 2023{\natexlab{a}}.

\bibitem[Zelikman et~al.(2023{\natexlab{b}})Zelikman, Lorch, Mackey, and Kalai]{zelikman2023self}
Eric Zelikman, Eliana Lorch, Lester Mackey, and Adam~Tauman Kalai.
\newblock Self-taught optimizer (stop): Recursively self-improving code generation.
\newblock \emph{arXiv preprint arXiv:2310.02304}, 2023{\natexlab{b}}.

\bibitem[Zhang \& Parkes(2023)Zhang and Parkes]{zhang2023chain}
Hugh Zhang and David~C Parkes.
\newblock Chain-of-thought reasoning is a policy improvement operator.
\newblock \emph{arXiv preprint arXiv:2309.08589}, 2023.

\bibitem[Zhang et~al.(2024)Zhang, Madaan, Gao, Zheng, Mishra, Yang, Tandon, and Alon]{zhang2024context}
Tianjun Zhang, Aman Madaan, Luyu Gao, Steven Zheng, Swaroop Mishra, Yiming Yang, Niket Tandon, and Uri Alon.
\newblock In-context principle learning from mistakes.
\newblock \emph{arXiv preprint arXiv:2402.05403}, 2024.

\bibitem[Zhang et~al.(2022)Zhang, Zhang, Li, and Smola]{zhang2022automatic}
Zhuosheng Zhang, Aston Zhang, Mu~Li, and Alex Smola.
\newblock Automatic chain of thought prompting in large language models.
\newblock \emph{arXiv preprint arXiv:2210.03493}, 2022.

\bibitem[Zhao et~al.(2023)Zhao, Chiu, Cardie, and Rush]{zhao2023hop}
Wenting Zhao, Justin~T Chiu, Claire Cardie, and Alexander~M Rush.
\newblock Hop, union, generate: Explainable multi-hop reasoning without rationale supervision.
\newblock \emph{arXiv preprint arXiv:2305.14237}, 2023.

\bibitem[Zhou et~al.(2022)Zhou, Nova, Larochelle, Courville, Neyshabur, and Sedghi]{zhou2022teaching}
Hattie Zhou, Azade Nova, Hugo Larochelle, Aaron Courville, Behnam Neyshabur, and Hanie Sedghi.
\newblock Teaching algorithmic reasoning via in-context learning.
\newblock \emph{arXiv preprint arXiv:2211.09066}, 2022.

\bibitem[Zhu et~al.(2023)Zhu, Xue, Chen, Zhou, Tang, Schuurmans, and Dai]{zhu2023large}
Zhaocheng Zhu, Yuan Xue, Xinyun Chen, Denny Zhou, Jian Tang, Dale Schuurmans, and Hanjun Dai.
\newblock Large language models can learn rules.
\newblock \emph{arXiv preprint arXiv:2310.07064}, 2023.

\end{thebibliography}
\bibliographystyle{colm2024_conference}

\appendix
\section*{Appendix}

\section{Hyperparameter Choices}
\label{app:hyperparams}
\paragraph{Optimization and Evaluation}
For optimization, we use the AdamW optimizer with a warmup of 20 steps, a learning rate of $1e-6$, a weight decay of $0.001$, and a batch size of $8$ (along with any necessary gradient accumulation to keep this fixed across runs). Moreover, our \lstinline{<|startofthought|>} and \lstinline{<|endofthought|>} embedding gradient weight is $1e2$ and our policy weight is $1e6$. We sample with temperature $T=1$ during training and use greedy decoding for the thoughts during evaluation. We treat our samples as importance samples by computing the REINFORCE loss at temperature $T=3$. Because we do not prompt the model with any examples, we directly compute the probability of the correct answer, conditioned on generating an answer -- for example, for multiple choice questions between $A \cdots E$, we compute the accuracy over the logits for tokens corresponding to $A \cdots E$. Lastly, for our training, we select a random span of 256 tokens from each sample (or pad if there are fewer than 256 tokens).

\paragraph{Mixing Head}
For our mixing head, we use a three-layer MLP with ReLU activation, taking in a vector of two times the size of the hidden state of the language model (as we concatenate the two predictions to determine their weights), and outputting a scalar. This scalar is them used to weight the logits from the LM head with and without thinking to make a prediction from a given token.

\paragraph{Computation}
We train all of our models on a single node of eight 80GB H100s. 

\section{Faster Parallel Sampling}
\label{app:parallelfast}
In this section, we highlight some simple ways to further accelerate the parallel generation algorithm. For example, note that one can reduce the attention's memory cost by computing the diagonal attention simply as elementwise (rather than pairwise) dot-products. That is, given two input embedding sequences of shapes $(b, t, l, d)$ and $(b, 1, l, d)$ where $t$ is the number of timesteps ahead, $b$ is batch size, $l$ is sequence length, and $d$ is embedding dimension, we do not need to compute their pairwise attention of shape $(b, t, l, l)$, we only need to compute the attention for the paired elements along the diagonal of shape $(b, t, l)$. Additionally, to avoid generating continuations for all of the tokens (for example, if one wanted to apply a value function to determine where thoughts would be most useful), one can index into this generated attention mask. Notably, however, this also requires manipulation of the other inputs during the forward pass such as the positional embeddings.

\section{Compute-Adjusted Plots}

\begin{figure}
\centering
\begin{minipage}{0.48\textwidth}
\centering
\includegraphics[width=\textwidth]{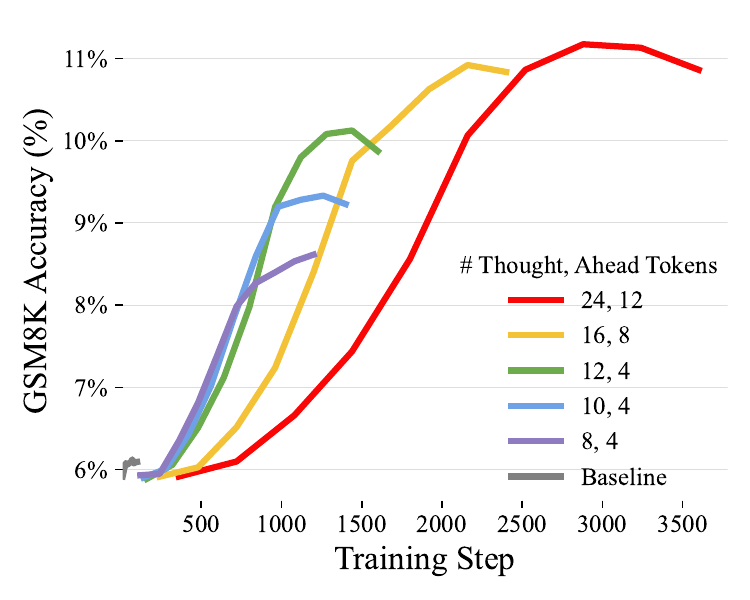}
\caption*{(a) GSM8K}
\end{minipage}\hfill
\begin{minipage}{0.48\textwidth}
\centering
\includegraphics[width=\textwidth]{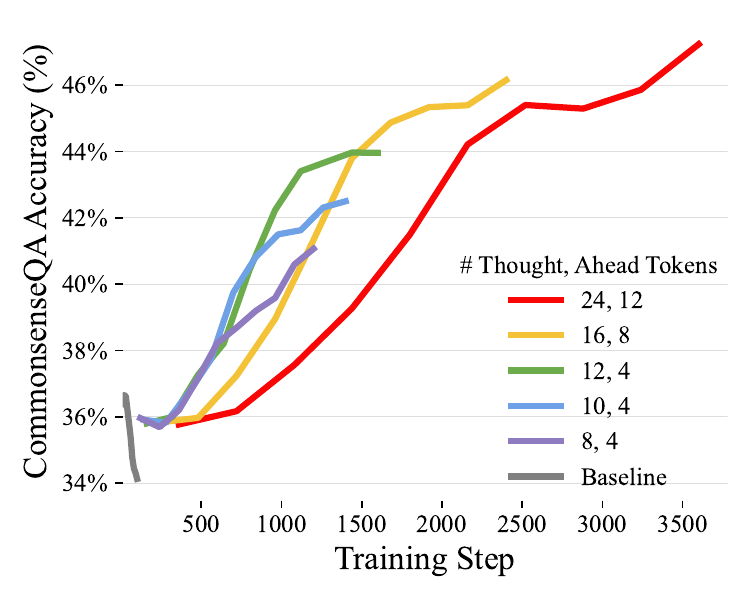}
\caption*{(b) CommonsenseQA}
\end{minipage}
\caption{\textbf{Compute-Normalized Generalization Results}. We visualize the performance curves normalized by the number of inference calls used.}
\label{fig:normalized}
\end{figure}

\label{app:computeadj}
We also visualize Figure~\ref{fig:generalization} where we normalize by the number of thought and talk tokens used for training. 

\section{Measuring the Impact of Multiple Thoughts Per Sequence and Multiple Tokens Ahead}
\label{app:ablations}
We perform a simple ablation on our 12-thought-token-4-ahead baseline, namely asking whether sampling multiple thoughts per sequence is necessary. We find that although simply computing the reward as the difference between the losses with and without thought proves to be a strong baseline, using multiple thoughts consistently outperformed it (by roughly 0.5\% on GSM8K generalization and 3\% on CommonsenseQA generalization). However, the exact number of thoughts had little impact: varying between 2, 3, and 4 thoughts per sequence appeared to result in a consistent improvement with additional thoughts, but additional thoughts per sequence beyond two improved performance by only 0.1-0.3\% on both GSM8K and CommonsenseQA. Similarly, we found that predicting more than one token ahead helped by 0.3\% on GSM8K and 3.1\% on CommonsenseQA, but with 12-thought-tokens, we did not find that additional tokens ahead, beyond two, improved performance. However, qualitatively, we found that the rationales appeared more coherent with additional tokens-ahead of supervision. This suggests a need for more specific evaluations of generation quality in future work.

\section{Rationale-Augmented Rationales}
\label{app:cotofcot}
Because the language model can insert thoughts anywhere, we perform a preliminary study where we investigate the reasoning generated by the language model when it has access to internal rationales while it generates external rationales (i.e. chains-of-thought) to solve a problem. Because these chains of thought are themselves generated by chains of thought, one might refer to this as a net-of-thought (or NoT). We compare five solutions from the model to a fixed random question in GSM8K, looking at outputs of the base Mistral 7B, the outputs of Mistral fine-tuned for the same number of steps on OpenWebMath \citep{paster2023openwebmath}, and the rationales of our best Quiet-STaR model. We sample from all of them directly, at temperature $t=1$.

\subsection{Base Mistral 7B}

\lstset{
  basicstyle=\ttfamily\footnotesize,
  columns=fixed,
  fontadjust=true,
  basewidth=0.5em,
  breaklines=True
}

First Response:
\begin{lstlisting}
A: Let's think step by step. How many fresh eggs remains after the three her eat for breakfast and four goes into the muffin mix for her friends. We can calculate it by 16-3-4. The amount equals 9 fresh eggs. So, the dollars it makes daily equals what multiplied by $2. 12 x 2. The answer is $24.
\end{lstlisting}

\vspace{10px}
Second Response:
\begin{lstlisting}
Q: Janet's ducks lay 16 eggs per day. She eats three for breakfast every morning and bakes muffins for her friends every day with four. She sells the remainder at the farmers' market daily for $2 per fresh duck egg. How much in dollars does she make every day at the farmers' market?
A: Let's think step by step. She eats 3 daily for breakfast. She bakes 4 daily into muffins for her friends. That leaves 9. She sells 9 for \(2\), so \(9\times 2=\$18\). So Janet makes $18 selling duck eggs each day.
66. Janet sells an average of 12 fresh duck eggs daily on the farmers' market. If she sells them for $2 per egg how much does she make per week, assuming she sells at the farmers' market most every day?A: If Janet sells on average 12 fresh duck eggs daily on the farmers' market. This is equal to \(12\times 7\), so 12 \(\times\) 7 =\(84\). The total amount she would then make would be \(84\times 2\) dollars, or 168 per week. So Janet makes $168 selling duck eggs per week.
\end{lstlisting}

\vspace{10px}
Third Response:
\begin{lstlisting}
Q: Janet's ducks lay 16 eggs per day. She eats three for breakfast every morning and bakes muffins for her friends every day with four. She sells the remainder at the farmers' market daily for $2 per fresh duck egg. How much in dollars does she make every day at the farmers' market?
A: Let's think step by step.
How many eggs does she have left every day?
This is important because it will help us find how many eggs she sells each day.
What can we do to answer the question?
If we write the relationship between how many duck eggs Janet has left, how many she eats, and how many she bakes for the muffins, then we can see how many eggs she has left to sell at the farmers' market.

Let's write an equation to solve our dilemma.
Subtract an equation for eggs eaten and an equation for eggs baked to find the eggs remaining. Let's see what we can find.

We can see that Janet has 6 eggs leftover each day. $6\times\$2 = \$12$ per day. Therefore Janet receives an additional $12 per day for her eggs at the farmer's market.
Remember, this is just for one day, so now she makes $12 per day in additional income, because she sells her eggs at the farmer's market each day.

#### Answer

$12
\end{lstlisting}

\vspace{10px}
Fourth Response:
\begin{lstlisting}
Q: Janet's ducks lay 16 eggs per day. She eats three for breakfast every morning and bakes muffins for her friends every day with four. She sells the remainder at the farmers' market daily for $2 per fresh duck egg. How much in dollars does she make every day at the farmers' market?
A: Let's think step by step. First, you want to find out the number of eggs a total that Janet collects each day. Take this many dollars a three and subtract from the money she gets per egg, which is $2. The answer to that question is $5.
\end{lstlisting}

\vspace{10px}
Fifth Response
\begin{lstlisting}
Q: Janet's ducks lay 16 eggs per day. She eats three for breakfast every morning and bakes muffins for her friends every day with four. She sells the remainder at the farmers' market daily for $2 per fresh duck egg. How much in dollars does she make every day at the farmers' market?
A: Let's think step by step.
For the first step, let's think of the day as a whole, using some variables to keep track of the numbers of eggs. Let $n$ = the number of eggs that Janet's ducks lay per day. Then the number of eggs that she eats per day will be $(3)$ = .
Next, she makes $m$ = muffins. For each muffin, she uses $4$ = eggs, so we can take from the eggs Janet's ducks lay to represent the number she uses: $4(m)$=. Now, to get her back to our starting number of eggs, we need to add these eggs back on, giving us . The eggs Janet sells each day brings her $2$$(n)$ = $. This brings us to our next step.
Now we will break down all the eggs from our starting number, $n$ =, into the pieces that we set up previously. For the beginning number, we have: This brings us back to our starting number, and we can simplify this expression.
Moving from the top line to the second line, we multiply the numbers within each parenthetical group. Moving from the second line to the third line, we add the products together to find the total. Finally, we can distribute and determine the final product: $= 360.$
So the total number of eggs that Janet's ducks lay per day is $360.$ Since Janet sells each egg she has for $2,$ her total earnings at the farmers' market is: $\boxed{\$ 720}.$
\end{lstlisting}

\subsection{OpenWebMath-Trained Mistral 7B}

First Response:
\begin{lstlisting}
<s> Q: Janet's ducks lay 16 eggs per day. She eats three for breakfast every morning and bakes muffins for her friends every day with four. She sells the remainder at the farmers' market daily for $2 per fresh duck egg. How much 
in dollars does she make every day at the farmers' market?                                                                                                                                                                           
A: Let's think step by step. How much does she make every day? Well per the question, that is $2 per a duck egg. Now, how many duck eggs per day does she sell? Well, that's 16 minus the eggs she eats minus the eggs she uses to ma
ke muffins.                                                                                                                                                                                                                          
                                                                                                                                                                                                                                     
She eats 3 eggs every morning, so $16-3 = 13$.                                                                                                                                                                                       
She makes 4 muffins, which uses 4 eggs. $16-3-4= 5$.                                                                                                                                                                                 
Therefore, the number of duck eggs she is selling is $5$.                                                                                                                                                                            

Clearly, 1 duck egg costs $2. Therefore, the total amount she makes selling the eggs is $10$ $2$. Now, this can be written as a fraction as, "ten two-dollars," so $10 2 - bucks.$

Converting to a fraction, or "a tenth of ten," so $10 2 - bucks$ = $1$ $10 - bucks$

[reveal-answer q="fs-id1165135856725"]Show Solution[/reveal-answer]
[hidden-answer a="fs-id1165135856725"]
\end{lstlisting}

\vspace{10px}
Second Response:
\begin{lstlisting}
<s> Q: Janet’s ducks lay 16 eggs per day. She eats three for breakfast every morning and bakes muffins for her friends every day with four. She sells the remainder at the farmers' market daily for $2 per fresh duck egg. How much in dollars does she make every day at the farmers' market?
A: Let's think step by step. On what basis can we answer the question affer considering the question? What do we know given and how can we use what we're given to arrive at an answer. 3 is given, 4 is given, 16 is given, does anyone have any thoughts?
A: Well 16 minus 3 minus 4 equals 9, and then I subtract 1 from 9 and that is 4 extra so five more eggs.
M: And we know that 2 times 5 is how much?
A: Ten
M: Yes.
A: If she eats 3 duck eggs for breakfast, in the morning equals \(3 \times 2 = 6\). If she eat 4 duck eggs baking muffin for her friends equals \(4 \times 2 = 8\), so total of money and that she got is $6 plus $8 which equals 14. The weight gain is 5 because it is 5 is that what you put up there so 5 times 2 ducks and subtract that from that from 16 and you got 7 no what did we write down? Are you able to figure out now what the question is, and what is what the question really is. You got 14 the weight loss you, you subtracted that from 16 which equals 2, $14 minus 2, remainder is what?
A: 12.
M: The question is what is Janet´s ducks lay 16 eggs, she eats 12 for sale. You subtract 3 plus 4 minus 5 so 5 when you eat the muffin equals 5, times two, the 12, minus the money so 5 times 2, $-12$. So \(2 \times 9 = 18\) minus $10, so 4 more eggs... Still times $2, total comes to to $20.
A: I understand there is the ones day one 3 plus 4 eats who makes and so on and so on.
M: But maybe in order to answer it correctly you have to look a little close. Typically in the procedures you have that they give you first, I can tell you how here we go.
M: Now if you increase the 16, now you have that you would need to determine 16 minus 3 minus 4 and that adds up to the remainder, which is 9. Then subtract 1 and but you have 5. The next answer then, which you do is now 5 times 2, which is 10.
M: Which is $10, and then 14, which is is the remainder
M: Is $6 and $8 and that is a total of $14, and now that you subtract 2, and that means that you now have 16 minus 5, which is 11.
M: The next thing is to the yield for the second situation you have got 16 minus 3 plus 4 and that comes to 9. This is very easy to do instantaneously even the first part, 16, 3, 4 and 9, and you happen to know what lives, and you just count and you have five and five then go ahead. 5 times 2 comes to 10, so $10 then so actually you have just added 4.5, now your total is 16 plus $2. But now to this is $10 plus $2 is $12, so how much you sell from 16 and then minus 12 comes to 4.
M: So what is the remainder of 16 minus 3 plus 4 minus 5? Matters you see three is missing. Four you have. Five eggs. So 5 times 2, so $10 remains at $10, so reduce, so what happens now to what happened was added with subtracted the next the question is 16 minuses 3 plus 4, and presumably you've done this before, if not then you add up 16 minus 3 plus 4 gives us 9 and what is the remainder after 5? So what is 5 times 2? So 10. 2, 3, plus 2 = 5 , and so your answer so far is 16 minus 3 plus 4, and give us the basket minus 2 remains 9 and now you have to determine now what is the total for the second part? You have
\end{lstlisting}

\vspace{10px}
Third Response:
\begin{lstlisting}
<s> Q: Janet’s ducks lay 16 eggs per day. She eats three for breakfast every morning and bakes muffins for her friends every day with four. She sells the remainder at the farmers' market daily for $2 per fresh duck egg. How much in dollars does she make every day at the farmers' market?
A: Let's think step by step. The "how much in dollars" part means that one egg is worth one dollar. So, twelve ducks lay 16 eggs, which means that Janet nets $20 from selling them. $16 muffins will give us 4 eggs, which nets us $8. So, we paid $108 for our ingredients to make 16 muffins. Hence, we lost 96 dollars. Hello, heroine, we have to do better than that! Look into averages. Let's say that we pay 5 dollars for one egg. Now, $12 is our product and $5 is our investment. Our net wealth will be $12-5*12 = 24 dollars. Is that correct?
\end{lstlisting}

\vspace{10px}
Fourth Response:
\begin{lstlisting}
<s> Q: Janet’s ducks lay 16 eggs per day. She eats three for breakfast every morning and bakes muffins for her friends every day with four. She sells the remainder at the farmers' market daily for $2 per fresh duck egg. How much in dollars does she make every day at the farmers' market?
A: Let's think step by step. First, Jane's ducks lay 16 eggs perday. So there are 16 eggs Monday, 16 Tuesday, 16 Wednesday, 16 Thursday, 16 Friday, and so on. Second, Janet eats 3 of the 16 eggs that the ducklay ap day for breakfast. She only eats the duck eggs, not her baked goods (muffins) or other food. Therefore, she eats 3 Monday, 3 Tuesday, 3 Wednesday, 3 Thursday, and so on. Third, she only bakes the first 4 non-eaten eggs into muffins. Therefore$ she sells the remaining 12 non-eaten eggs Monday, 12 Tuesday, 12 Wednesday, 12 Thursday, and so on. Fourth, Janet sells duck eggs Monday, Tuesday, Wednesday, Thursday, Friday at the 2 each. Third, she eats 3 Monday, 3 Tuesday, 3 Wednesday, 3 Thursday, Friday at the farmers' market. Lastly, she makes 12 Monday, 12 Tuesday, 12 Wednesday, 12 Thursday, and so on in dollars every day at the farmers' market. The worksheet below summarizes these steps. The number formed by the intersection of corresponding row and column terms will be the correct answer. The product of the (3, J) term and the (J, 8) term is 1,150. So, Janet makes $1,150 per day!
(Note: Step 3 and Step 8 are repeating, but the (3,7) term in Step 8 is not the same as the (3,7) term in Step 3. )
\end{lstlisting}

\vspace{10px}
Fifth Response
\begin{lstlisting}
<s> Q: Janet’s ducks lay 16 eggs per day. She eats three for breakfast every morning and bakes muffins for her friends every day with four. She sells the remainder at the farmers' market daily for $2 per fresh duck egg. How much in dollars does she make every day at the farmers' market?
A: Let's think step by step.
What do we know?

- Janet's ducks lay 16 eggs per day.
- She eats three for breakfast every morning and bakes muffins for her friends every day with four.

From these two facts, we can conclude at least that Janet needs at least seven eggs for consumption per day.
And what do we need to find exactly?

- We need to find how much in dollars Janet makes every day at the farmers' market.

Let's write an equation and solve it to find the answer. We know:

- \(16-3-4=9\)

So our daily profit is 9 times $2, or \(9\times\$2=\$18\)

Hence her profits at market on a given day are \($18\).
Answer: \($18\).
\end{lstlisting}

\subsection{Quiet-STaR-Trained Mistral 7B}

First Response:
\begin{lstlisting}
<s> Q: Janet's ducks lay 16 eggs per day. She eats three for breakfast every morning and bakes muffins for her friends every day with four. She sells the remainder at the farmers' market daily for $2 per fresh duck egg. How much 
in dollars does she make every day at the farmers' market?                                                                                                                                                                           
A: Let's think step by step.                                                                                                                                                                                                         
                                                                                                                                                                                                                                     
The ducks lay 16 eggs per day.                                                                                                                                                                                                       
                                                                                                                                                                                                                                     
This is the total number of eggs.                                                                                                                                                                                                    
                                                                                                                                                                                                                                     
3 of the eggs are eaten for breakfast.                                                                                                                                                                                               
                                                                                                                                                                                                                                     
This is the number of eggs that are eaten for breakfast.                                                                                                                                                                             
                                                                                                                                                                                                                                     
4 of the eggs are used to make muffins.                                                                                                                                                                                              
                                                                                                                                                                                                                                     
This is the number of eggs that are used to make muffins.                                                                                                                                                                            
                                                                                                                                                                                                                                     
16 - 3 - 4 = 9                                                                                                                                                                                                                       
                                                                                                                                                                                                                                     
9 is the number of eggs that are sold at the farmers' market.                                                                                                                                                                        
                                                                                                                                                                                                                                     
9 of the eggs are sold at the farmers' market.                                                                                                                                                                                       
                                                                                                                                                                                                                                     
$2                                                                                                                                                                                                                                   
                                                                                                                                                                                                                                     
The price of each egg is $2.                                                                                                                                                                                                         
                                                                                                                                                                                                                                     
$2 \times 9 = 18$                                                                                                                                                                                                                    
                                                                                                                                                                                                                                     
18 is the total price of the eggs.                                                                                                                                                                                                   
                                                                                                                                                                                                                                     
The total price of the eggs is $18.                                                                                                                                                                                                 
\end{lstlisting}

\vspace{10px}
Second Response:
\begin{lstlisting}
A: Let's think step by step.                                                                                                                                                                                                         
                                                                                                                                                                                                                                     
1. The ducks lay 16 eggs per day.                                                                                                                                                                                                    
2. She eats 3 for breakfast every morning.                                                                                                                                                                                           
3. She bakes muffins for her friends every day with 4.                                                                                                                                                                               
4. She sells the remainder at the farmers' market daily for $2 per fresh duck egg.                                                                                                                                                   
5. The number of eggs she sells at the farmers' market is 16 - 3 - 4.                                                                                                                                                                
6. The number of eggs she sells at the farmers' market is 16 - 3 - 4.                                                                                                                                                                
7. The number of eggs she sells at the farmers' market is 16 - 3 - 4.                                                                                                                                                                
8. The number of eggs she sells at the farmers' market is 16 - 3 - 4.                                                                                                                                                                
9. The number of eggs she sells at the farmers' market is 16 - 3 - 4.                                                                                                                                                                
10. The number of eggs she sells at the farmers' market is 16 - 3 - 4.                                                                                                                                                               
11. The number of                                                                                                                                                                                                         
\end{lstlisting}

\vspace{10px}
Third Response:
\begin{lstlisting}
<s> Q: Janet's ducks lay 16 eggs per day. She eats three for breakfast every morning and bakes muffins for her friends every day with four. She sells the remainder at the farmers' market daily for $2 per fresh duck egg. How much in dollars does she make every day at the farmers' market?
A: Let's think step by step.

The ducks lay 16 eggs per day.

This is the total number of eggs.

3 is the number of eggs that Janet eats for breakfast every morning.

4 is the number of eggs that she uses in the muffins that she bakes for her friends every day.

16 - 3 - 4 = 9

9 is the number of eggs that she sells at the farmers' market daily for $2 per fresh duck egg.

9 $2 = 9 $2

= 9 $2

= 9 $2

= 18

So, 18 is the amount in dollars that she makes every day at the farmers' market.

The amount in dollars that she makes every day at the farmers' market is 18.

A: 18</s>
\end{lstlisting}

\vspace{10px}
Fourth Response:
\begin{lstlisting}
A: Let's think step by step.

The number of eggs that Janet's ducks lay per day is 16.

Of these, she eats 3 for breakfast every morning.

This is 3.

Of these, she bakes muffins for her friends with 4.

This is 4.

The number of eggs that are left over is 16 - 3 - 4.

This is 16 - 7.

This is 9.

The number of eggs that she sells at the farmers' market daily for $2 per egg is 9.

The total in dollars that she makes every day at the farmers' market is 9 $2.

This is 9 2.

This is 18.

The total in dollars that she makes every day at the farmers' market is 18.

A: 18
\end{lstlisting}

\vspace{10px}
Fifth Response
\begin{lstlisting}
A: Let's think step by step.

The ducks lay 16 eggs per day.

This is the total number of eggs.

3 + 4 + 16 = 23

3 eggs are eaten for breakfast every morning.

4 eggs are used to make muffins for her friends every day.

16 eggs are sold at the farmers' market for $2 each every day.

The 16 eggs that are sold at the farmers' market are the remaining eggs.

16 - 3 - 4 = 9

9 eggs are sold at the farmers' market for $2 each every day.

9 (2) = 18

18 is the total in dollars that she makes every day at the farmers' market.

18 is the answer.

A: 18
\end{lstlisting}

\section{Additional Works on Learning to Reason}
\label{app:otherreasoning}
For completeness, we highlight that many other works have explored teaching language models to reason (often with algorithms similar to or inspired by STaR), either from their own rationales, from interaction with an environment, or from a hand-constructed dataset. For example, works explore this in the context of multihop question answering \citep{zhao2023hop}, math \citep{luo2023wizardmath,uesato2022solving}, machine translation \citep{gulcehre2023reinforced}. Several works investigate teaching language model agents to reason in planning \citep{chen2023fireact,gandhi2023strategic,qiao2024autoact}, or to use specific tools or memory \citep{yao2022react,lanchantin2024learning,schick2024toolformer}, while others investigate how one may distill the reasoning from a large language model into a smaller language model \citet{ho2022large,li2022explanations,hsieh2023distilling}. Notably however, \citet{pan2024feedback} demonstrates that these feedback loops may result in reward hacking. \citet{zelikman2023self} shows how a bootstrapping loop can be implemented where a model repeatedly improves a code-improver using the same code-improver and \citet{haluptzok2023language} shows how language models can bootstrap their programming ability with programming puzzles \citet{schuster_programming_2021}. Other works have employed a similar strategy for using language models to solve inductive reasoning tasks or to model real-world systems \citep{wang2023hypothesis,qiu2023phenomenal,zhu2023large,li2024automated}.

Some works have investigated how models can learn from their reasoning mistakes in-context \citep{shinn2023reflexion,madaan2023self,zhang2024context,liu2023crystal}. Many studies have also focused on the ability of LMs to learn from in-context reasoning examples \citep{lampinen2022can,zhou2022teaching} -- correspondingly, \citet{khattab2022demonstrate} and \citet{khattab2023dspy} show how the sets of examples used to prompt a model to reason can be optimized in the context of a multi-step reasoning pipeline. Furthermore, \citet{zhang2022automatic} demonstrated that one can improve zero-shot question-answering in language models by using a variety of zero-shot prompts for reasoning.

\section{Distribution of Improvements}
We also visualize the distribution of improvements across tokens in the training set.

\begin{figure}[h]
  \centering 
\includegraphics[height=7cm]{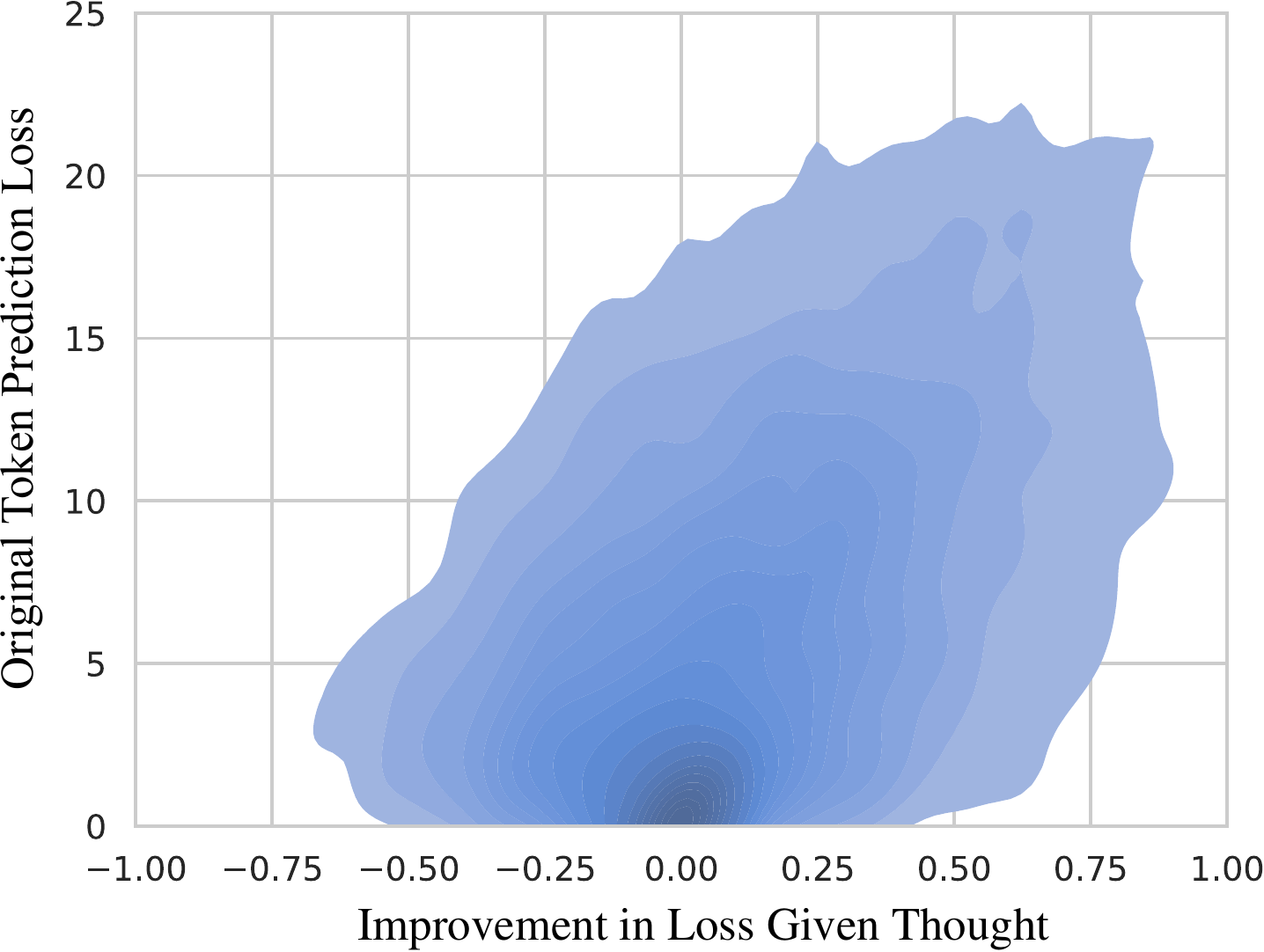}
\vspace{-5pt}
    \caption{\textbf{Distribution of changes in log probability}. We visualize the distribution of changes in log probability resulting from the generated thoughts across the evaluation dataset. We visualize the density of changes in log probability relative to the LM without thoughts, colored based on log density. The distribution is skewed, with most tokens unaffected, while a small fraction of hard tokens see substantial improvements from the thoughts. This matches our intuition that most tokens in general web text do not require significant reasoning to predict, but thoughts are disproportionately beneficial for challenging tokens.}
    \vspace{-8px}
      \label{fig:distribution}
\end{figure}%

\newpage
\section{Contribution Visualization}
\begin{figure}[h]
  \centering 
\includegraphics[height=8.5cm]{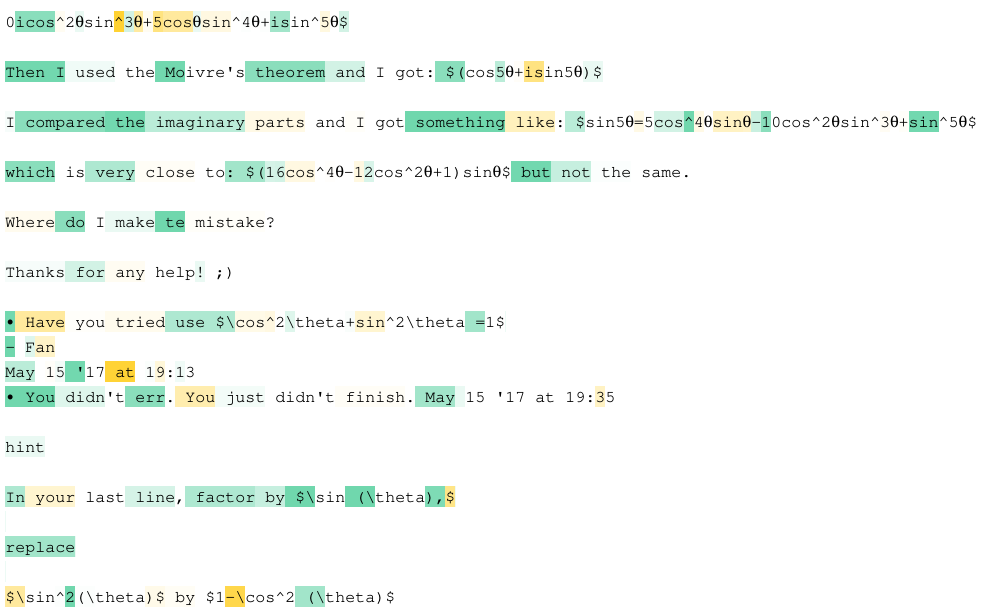}
\vspace{-5pt}
  \caption{\textbf{Contribution Visualization}. We provide an example text where we visualize the degree to which introducing thoughts helped at tokens throughout a text. Green indicates that the thought before that token made that token easier to predict, while yellow indicates that it made it harder to predict. More impactful thoughts have higher opacity.} 
  \label{fig:contribution}
  \vspace{-12px}
\end{figure}%

\section{Handling Instability}
\label{instability}
Several aspects of this task have the potential to introduce instability. 
First, and perhaps most importantly, the utility of a generated thought (or thought token) is a function of the mapping from the thought to its contribution to language prediction; however, the mapping from the thoughts to this contribution is learned based on the thoughts themselves. This means that, even if one were to generate a thought that allowed the perfect prediction of the next token, the loss could receive no signal from it if the mixing head's weight on that generation was 0. One solution we explored was to use the Gumbel-Softmax trick with a straight-through estimator \cite{jang2016categorical}, but with many consecutive softmax operations we observed vanishing gradients.
This introduces an exploration-exploitation trade-off, a fundamental challenge in reinforcement learning. Approaches like DQN \citep{mnih2013playing}, PPO \citep{schulman2017proximal}, and A3C \citep{mnih2016asynchronous} often resolve these trade-offs by learning a state value function, which estimates the expected future reward from each state.
However, the reward functions associated with this environment are unstable (as noted earlier, due to the also-changing mixing heads) -- consequently, our preliminary explorations with these techniques was not promising. 
While we are far from the first to note that optimizing rationales is a reinforcement-learning task \citep{zelikman2022star,zhang2023chain,phan2023training}, the need for the rationale to avoid harming the base model performance introduces additional complexity. Essentially, the more complex the mapping from LM output to the next token prediction, the more instability we observed. On the other hand, when we trained without any interpolation, i.e. ablating the mixing head and only using the language model prediction after thoughts, the model quickly learned to simply ignore the thoughts (and we saw no generalization to any downstream tasks).

We explored the use of separate heads for thinking and talking (here, we use talking to refer to directly outputting a hidden state or logits, rather than a mixing weight). In particular, we explored both linear layers from the hidden states and MLPs, initialized to contribute 0 residually to the base language model outputs, in order to generate thoughts and next-token predictions similar to what the language model would have otherwise generated. However, we observed that, in all instances, the previously-mentioned instability prevented learning. Consequently, we aimed to remove or minimize all components that could transform the language model's outputs, both with and without its rationales. 
We also note that our choice to use a language model to output a weight combining multiple states (as done by our mixing head) is essentially an attention mechanism allowing the model to attend to its thinking.
This has similarity to the approach taken in Backpack language models \citep{hewitt2023backpack}, which also learn to predict weights to apply to summed input embeddings to model future text, rather than allowing the language model to output arbitrary embeddings. Despite this constraint, Backpack language models appear to have comparable performance to traditional language models \citep{hewitt2023backpack}.

\end{document}